%% file: main.tex
\documentclass[fleqn,10pt]{wlscirep}
\usepackage[utf8]{inputenc}
\usepackage[T1]{fontenc}
\usepackage{bm}

\usepackage{booktabs}
\usepackage{multirow}
\usepackage{graphicx}

\usepackage{longtable}
\usepackage{caption}
\usepackage{xurl}
\usepackage{color}
\usepackage{multirow}

\usepackage{lineno}

\title{Native Intelligence Emerges from Large-Scale \\ Clinical Practice: A Retinal Foundation Model with Deployment Efficiency}

\author[1,\#]{Jia Guo}
\author[2,4,\#]{Jiawei Du}
\author[3,4]{Shengzhu Yang}
\author[2,4]{Shuai Lu}
\author[1]{Wenquan Cheng}
\author[5]{Kaiwen Zhang}
\author[1]{Yihua Sun}
\author[2]{Chuhong Yang}
\author[3,4,*]{Weihang Zhang}
\author[6,*]{Fang Chen}
\author[7,8]{Yilan Wu}
\author[8,9]{Lie Ju}
\author[1,10]{Guochen Ning}
\author[1]{Longfei Ma}
\author[11]{Huiping Yao}
\author[7]{Jinyuan Wang}
\author[12]{Peilun Shi}
\author[8,9,13]{Yukun Zhou}
\author[5]{Jie Xu}
\author[8,9]{Pearse A. Keane}
\author[4,5,14,*]{Hanruo Liu}
\author[1,6,*]{Hongen Liao}
\author[4,5,15,*]{Ningli Wang}
\author[2,4,*]{Huiqi Li}
\affil[1]{School of Biomedical Engineering, Tsinghua Medicine, Tsinghua University, Beijing, China}
\affil[2]{School of Information and Electronics, Beijing Institute of Technology, Beijing, China}
\affil[3]{School of Medical Technology, Beijing Institute of Technology, Beijing, China}
\affil[4]{Beijing Key Laboratory of Intelligent Diagnosis Technology and Equipment for Optic Nerve-Related Eye Diseases, Beijing, China}
\affil[5]{Beijing Tongren Hospital, Capital Medical University, Beijing, China}
\affil[6]{School of Biomedical Engineering, Shanghai Jiaotong University, Shanghai, China}
\affil[7]{Beijing Visual Science and Translational Eye Research Institute (BERI), Beijing Tsinghua Changgung Hospital Eye Center, School of Clinical Medicine, Tsinghua Medicine, Tsinghua University}
\affil[8]{Institute of Ophthalmology, University College London, London, UK}
\affil[9]{NIHR Biomedical Research Centre at Moorfields Eye Hospital NHS Foundation Trust, London, UK}
\affil[10]{School of Clinical Medicine, Tsinghua Medicine, Tsinghua University}
\affil[11]{Department of Ophthalmology, Ruijin Hospital, Shanghai Jiao Tong University School of Medicine, Shanghai, China}
\affil[12]{Department of Biomedical Engineering, The Chinese University of Hong Kong, Hong Kong SAR, China}
\affil[13]{UCL Hawkes Institute, University College London, London, UK}
\affil[14]{Henan Provincial People's Hospital, Henan Eye Hospital, Henan, China}
\affil[15]{Henan Academy of Innovations in Medical Science, Zhengzhou, Henan, China}
\affil[$\#$]{Equal contribution} \affil[*]{These authors jointly supervised this work}

\begin{abstract}
Current retinal foundation models remain constrained by curated research datasets that lack authentic clinical context, and require extensive task-specific optimization for each application, limiting their deployment efficiency in resource-constrained settings.
Here, we show that these barriers can be overcome by building clinical native intelligence directly from real-world medical practice. Our key insight is that large-scale telemedicine programs, where expert centers provide remote consultations across distributed facilities, represent a natural reservoir for learning clinical image interpretation. We present ReVision, a retinal foundation model that learns from the natural alignment between 485,980 color fundus photographs and their corresponding diagnostic reports, accumulated through a decade-long telemedicine program spanning 162 medical institutions across China.
 Through extensive evaluation across 27 ophthalmic benchmarks, we demonstrate that ReVison enables deployment efficiency with minimal local resources. Without any task-specific training, ReVision achieves zero-shot disease detection with an average AUROC of 0.946 across 12 public benchmarks and 0.952 on 3 independent clinical cohorts. When minimal adaptation is feasible, ReVision matches extensively fine-tuned alternatives while requiring orders of magnitude fewer trainable parameters and labeled examples. The learned representations also transfer effectively to new clinical sites, imaging domains, imaging modalities, and systemic health prediction tasks. In a prospective reader study with 33 ophthalmologists, ReVision's zero-shot assistance not only improved diagnostic accuracy by 14.8\% across all experience levels, but also revealed valuable insights into the real-world clinical potential of a deployment-efficient model. These results demonstrate that clinical native intelligence can be directly extracted from clinical archives without any further annotation to build medical AI systems suited to various resource-constrained settings.  
\end{abstract}

\begin{document}

\flushbottom
\maketitle
\thispagestyle{empty}


\input{introduction}

\input{results}

\input{discussion}

\input{method}

\bibliography{ref}

\newpage
\setcounter{figure}{0}   
\renewcommand{\figurename}{Extended Data Figure}

\begin{figure}[!t]
\setlength{\abovecaptionskip}{0mm}
\centerline{\includegraphics[width=\linewidth]{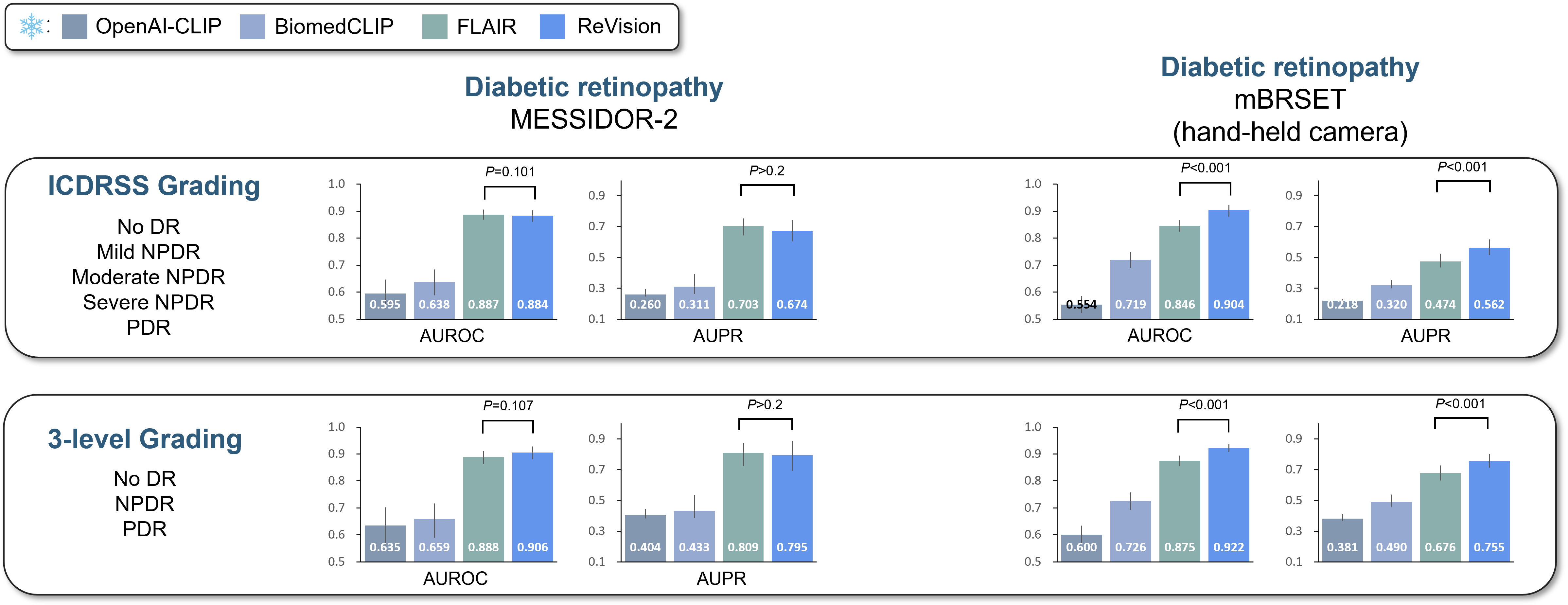}}
\caption{\textbf{Zero-shot performance on diabetic retinopathy (DR) grading.} FLAIR has been extensively pretrained on APTOS-2019 and IDRID for DR grading. Meanwhile, in the pretraining dataset (TM500k) of ReVision, only a small proportion of DR samples contain 5-level grading information in the reports, as ICDRSS standards are not a mandatory component in our clinical reports.  We employed non-parametric bootstrapping with 2,000 resamples to derive 95\% CIs. NPDR, non-proliferative DR. PDR, proliferative DR.
}
\label{fig_ex_drgrad}
\end{figure}

\begin{figure}[!t]
\setlength{\abovecaptionskip}{0mm}
\centerline{\includegraphics[width=\linewidth]{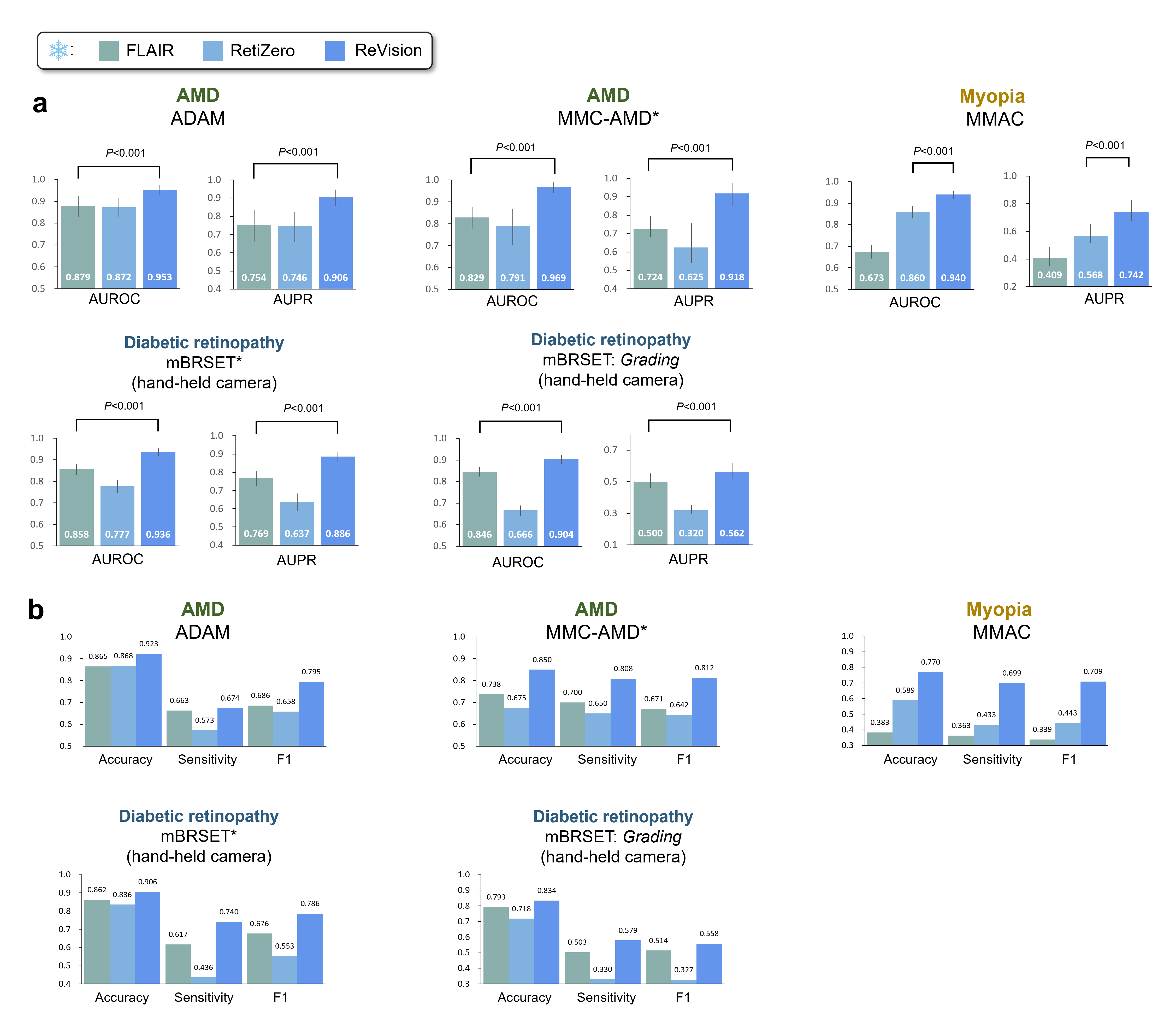}}
\caption{\textbf{Zero-shot performance on benchmarks that were not used in the pretraining of RetiZero and FLAIR.} \textbf{a,} AUROC and AUPR.  *: To build unambiguous zero-shot benchmarks, we trimmed the original datasets.  For MMC-AMD, we ignored the ambiguous `PCV' class as it can be categorized as wet-AMD. For mBRSET (DR grading in hand-held cameras), we employed both binary DR detection and 5-level DR grading. We employed non-parametric bootstrapping with 2,000 resamples to derive 95\% CIs. \textbf{b,} Accuracy, macro Sensitivity, and macro F1-score. To convert the probability values output by the model into categories, we optimized decision thresholds to maximize macro F1-score on the validation set of each dataset and applied them to the test set. 
}
\label{Fig_ex_4datazs}
\end{figure}

\begin{figure}[htbp!]
\setlength{\abovecaptionskip}{0mm}
\centerline{\includegraphics[width=0.85\linewidth]{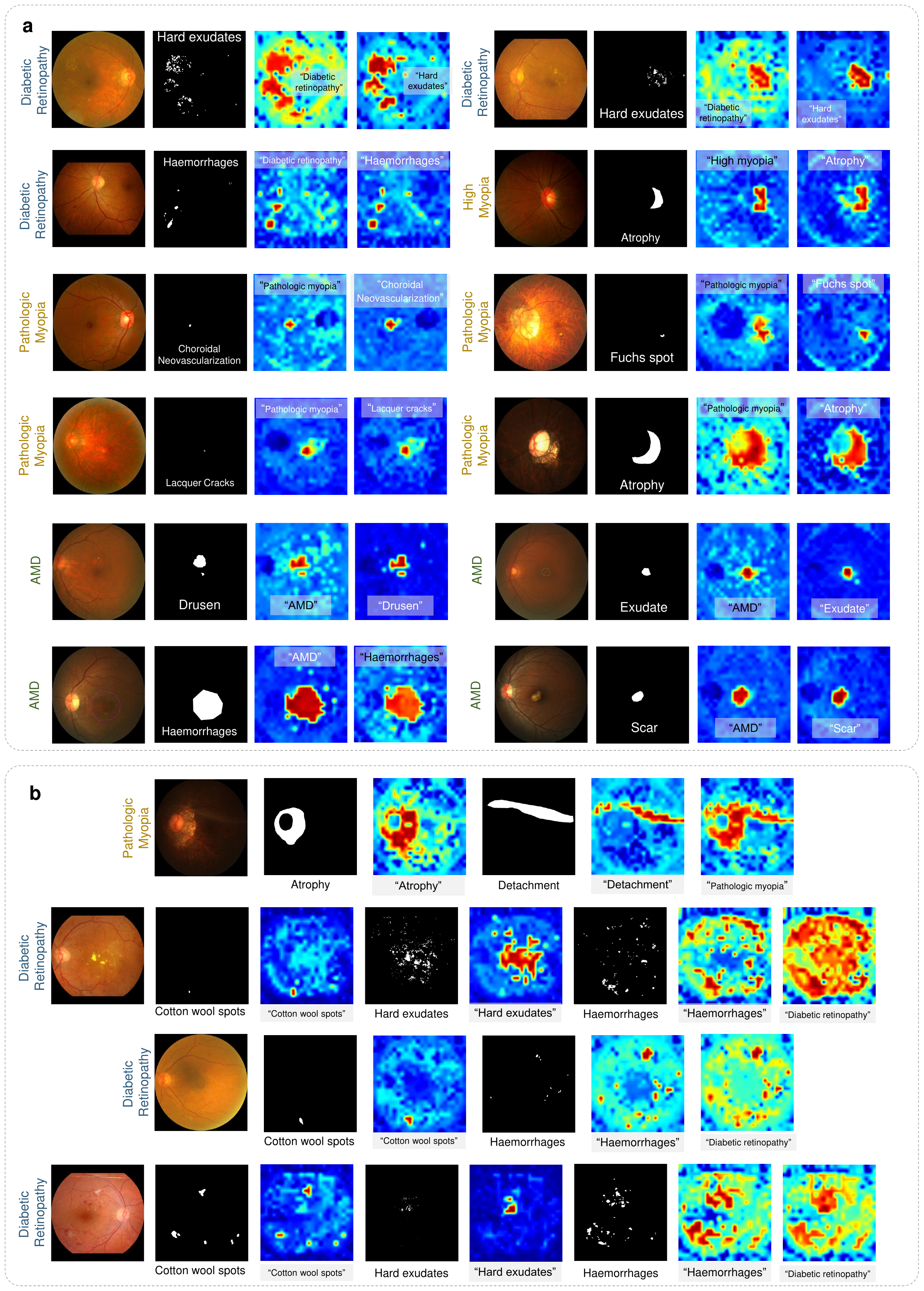}}
\caption{\textbf{Zero-shot lesion localization.} The heatmap (blue: low similarity --> red: high similarity) is generated by computing the cosine similarity between local image regions and the corresponding text prompt. ReVision is capable not only of localizing specific lesions in the image given their names, but also of identifying disease-associated pathological regions when provided with a disease-level prompt. As illustrated in \textbf{b}, we further evaluate ReVision on more complex scenarios where multiple lesion types co-occur within a single image. ReVision successfully highlights all lesion types that are semantically linked to the given disease prompt, demonstrating its ability to internalize both the anatomical structure and pathological characteristics of ophthalmic conditions.}
\label{Fig_ex_zs_loc}
\end{figure}

\begin{figure}[htbp!]
\setlength{\abovecaptionskip}{0mm}
\centerline{\includegraphics[width=\linewidth]{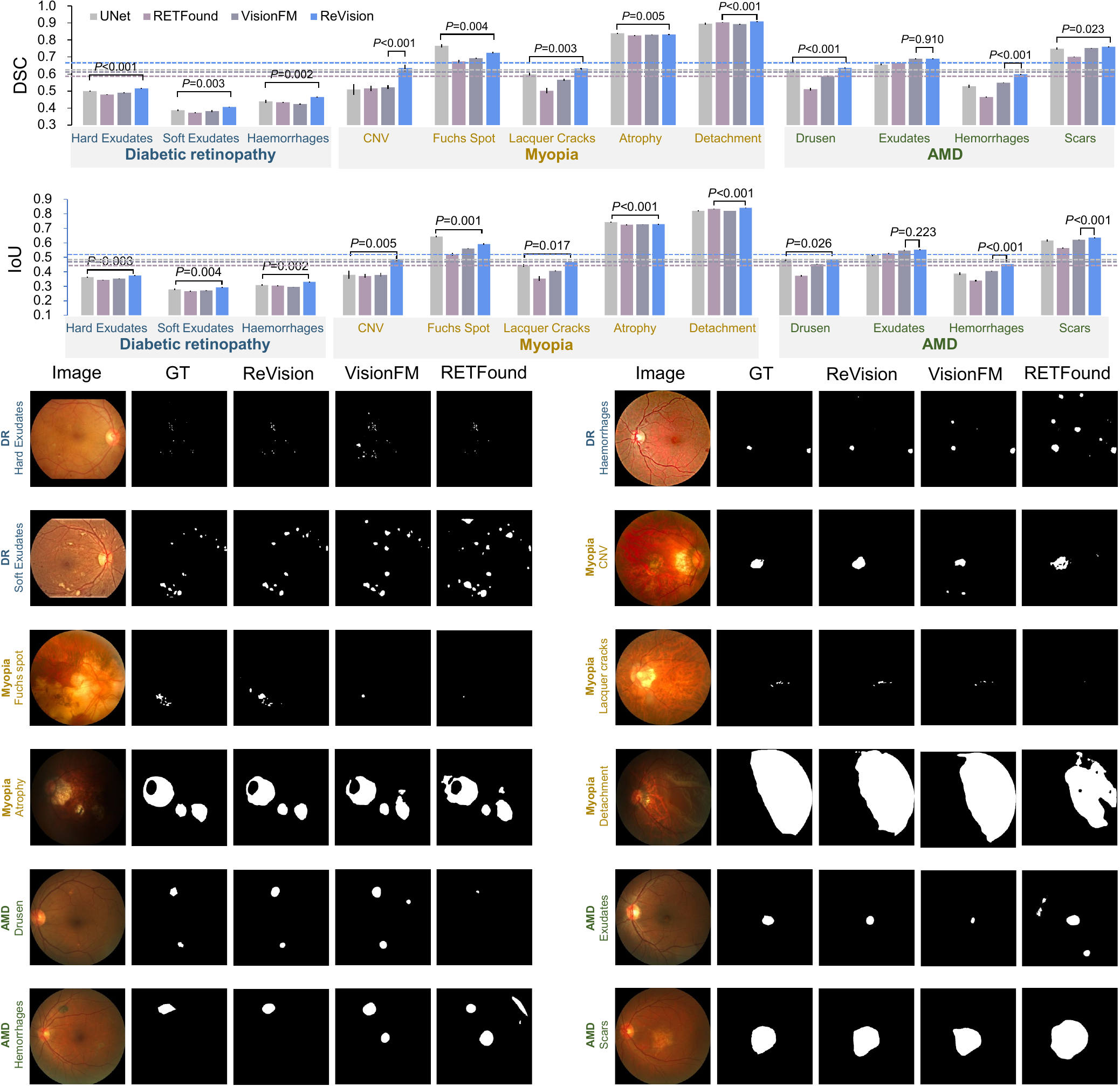}}
\caption{\textbf{Performance on supervised ocular disease lesion segmentation.} \textbf{a}, Segmentation performance measured by the IoU. For each task, all models were evaluated over five independent runs using different random seeds to account for variability. Results are reported as mean IoU across runs, with error bars indicating 95\% CI. Dashed lines represent each model’s mean performance averaged across all tasks. \textbf{b}, Qualitative comparison of segmentation results among ReVision, VisionFM, and RETFound.}
\label{Fig_ex_seg}
\end{figure}

\begin{figure}[!t]
\setlength{\abovecaptionskip}{0mm}
\centerline{\includegraphics[width=\linewidth]{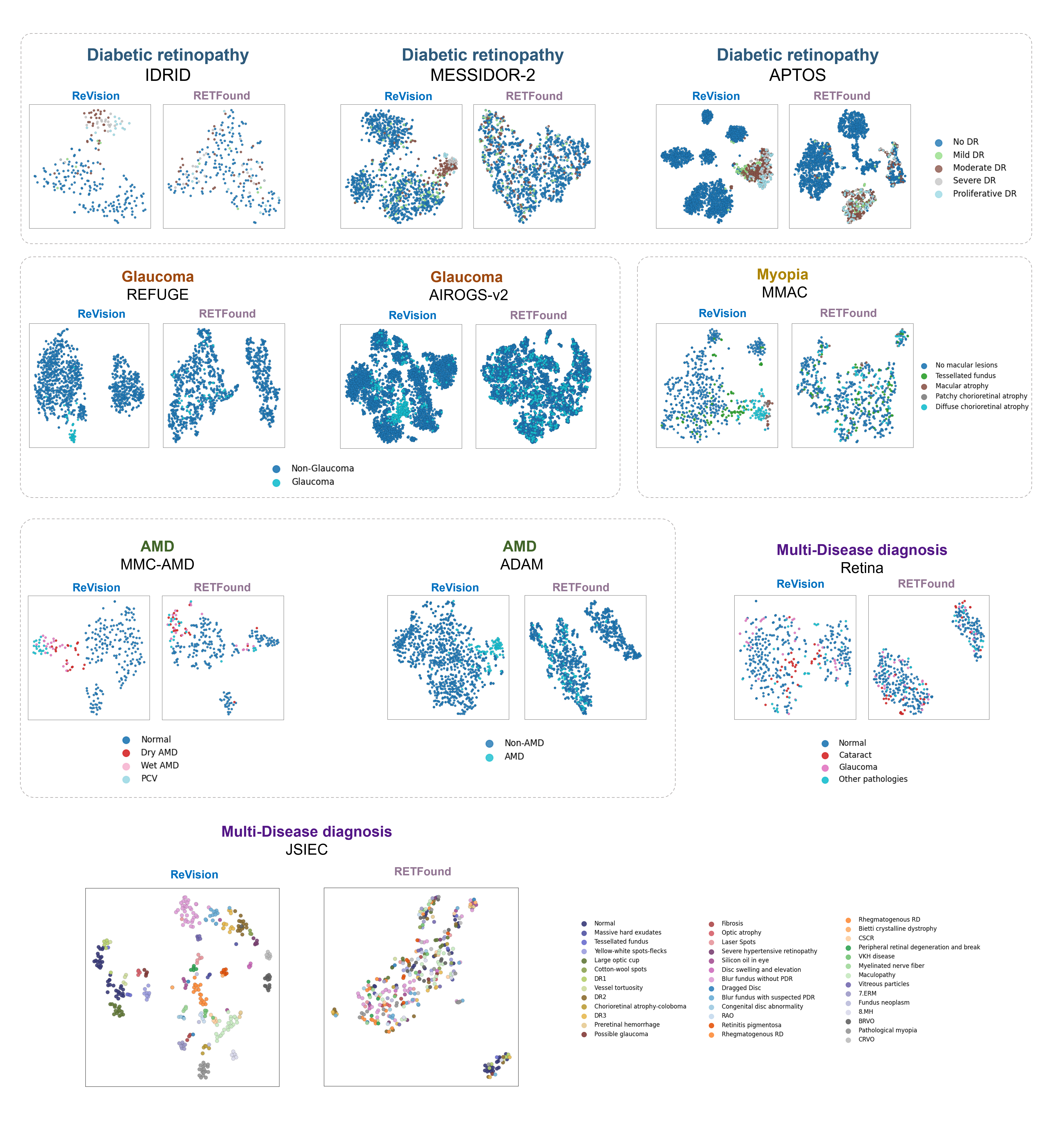}}
\caption{\textbf{t-SNE visualization of image embeddings encoded by ReVision and RETFound.}  Without any task-specific fine-tuning, ReVision’s pre-trained embeddings demonstrate distinct clustering patterns across different disease categories.
}
\label{Fig_ex_tnse}
\end{figure}

\begin{figure}[htbp]
\setlength{\abovecaptionskip}{0mm}
\centerline{\includegraphics[width=0.95\linewidth]{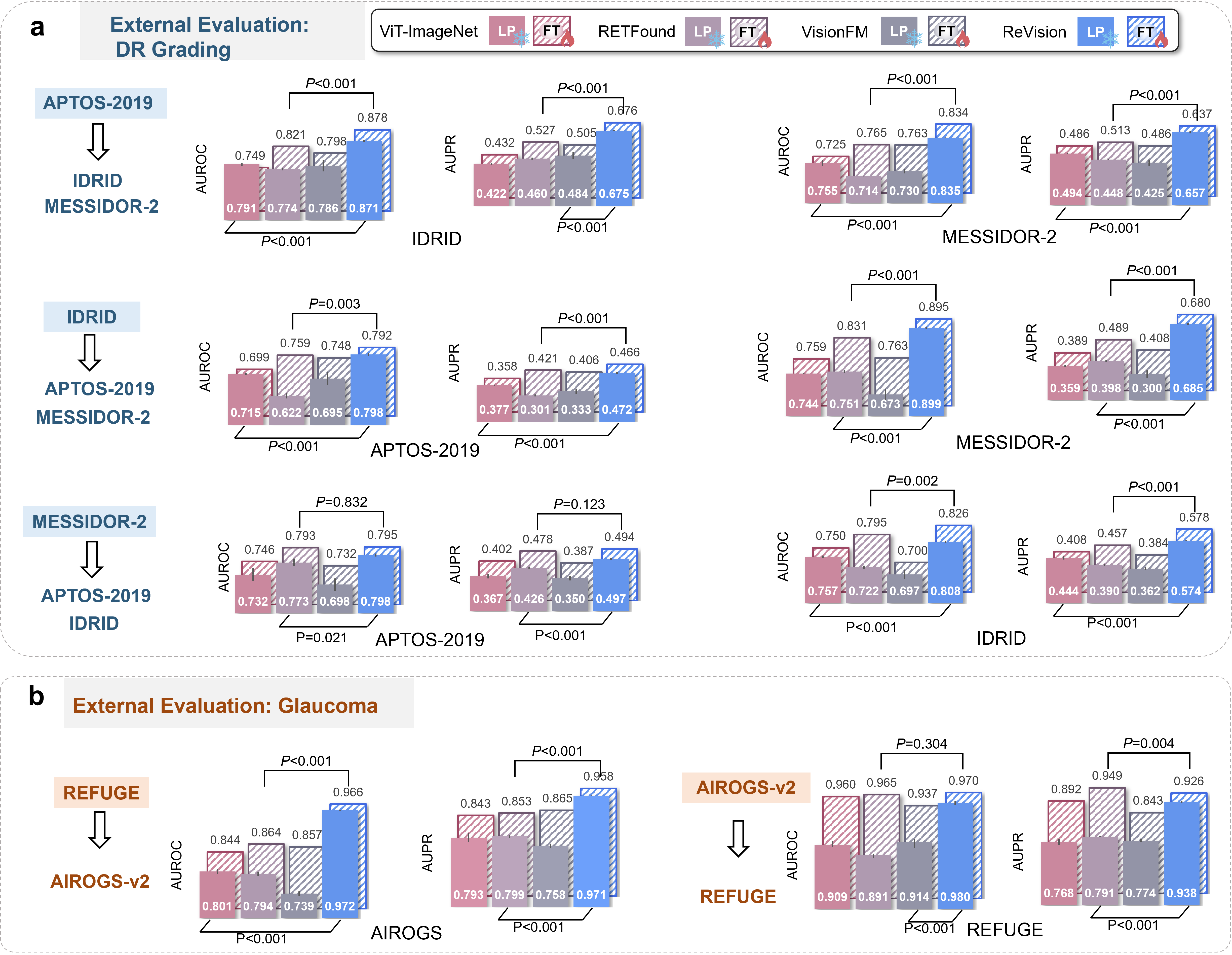}}
\caption{\textbf{Performance on external evaluation.} \textbf{a,} External evaluation results of diabetic retinopathy grading. The model tuning (LP or FT) and checkpoint selection were performed on one dataset, while external testing was conducted on others. \textbf{b,} External evaluation results of glaucoma diagnosis. Results are the mean across 5 independent experiments, and the error bars show 95\% CI. P-value is calculated with the two-sided t-test.}
\label{fig_ex_external}
\end{figure}

\begin{figure}[!t]
\setlength{\abovecaptionskip}{0mm}
\centerline{\includegraphics[width=0.95\linewidth]{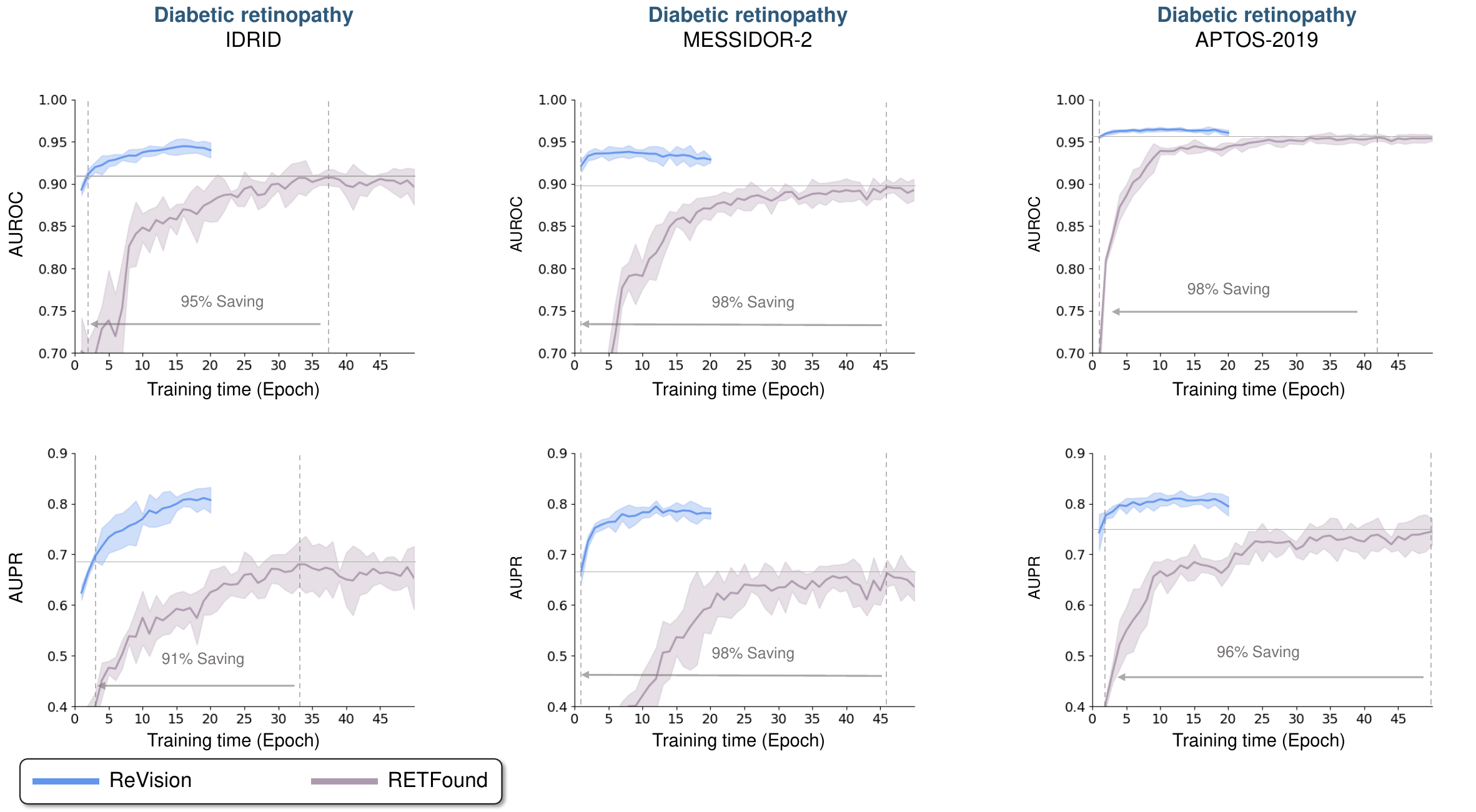}}
\caption{\textbf{Convergence efficiency during downstream training.} We tracked optimization dynamics of full fine-tuning by plotting validation performance after each epoch on IDRID, MESSIDOR-2, and APTOS-2019. ReVision achieved RETFound's peak performance in just 3 epochs. Center lines represent mean values across 5 runs, with colored bands indicating 95\% confidence intervals.
 }
\label{Fig_ex_converge}
\end{figure}

\begin{figure}[htbp]
\setlength{\abovecaptionskip}{0mm}
\centerline{\includegraphics[width=0.95\linewidth]{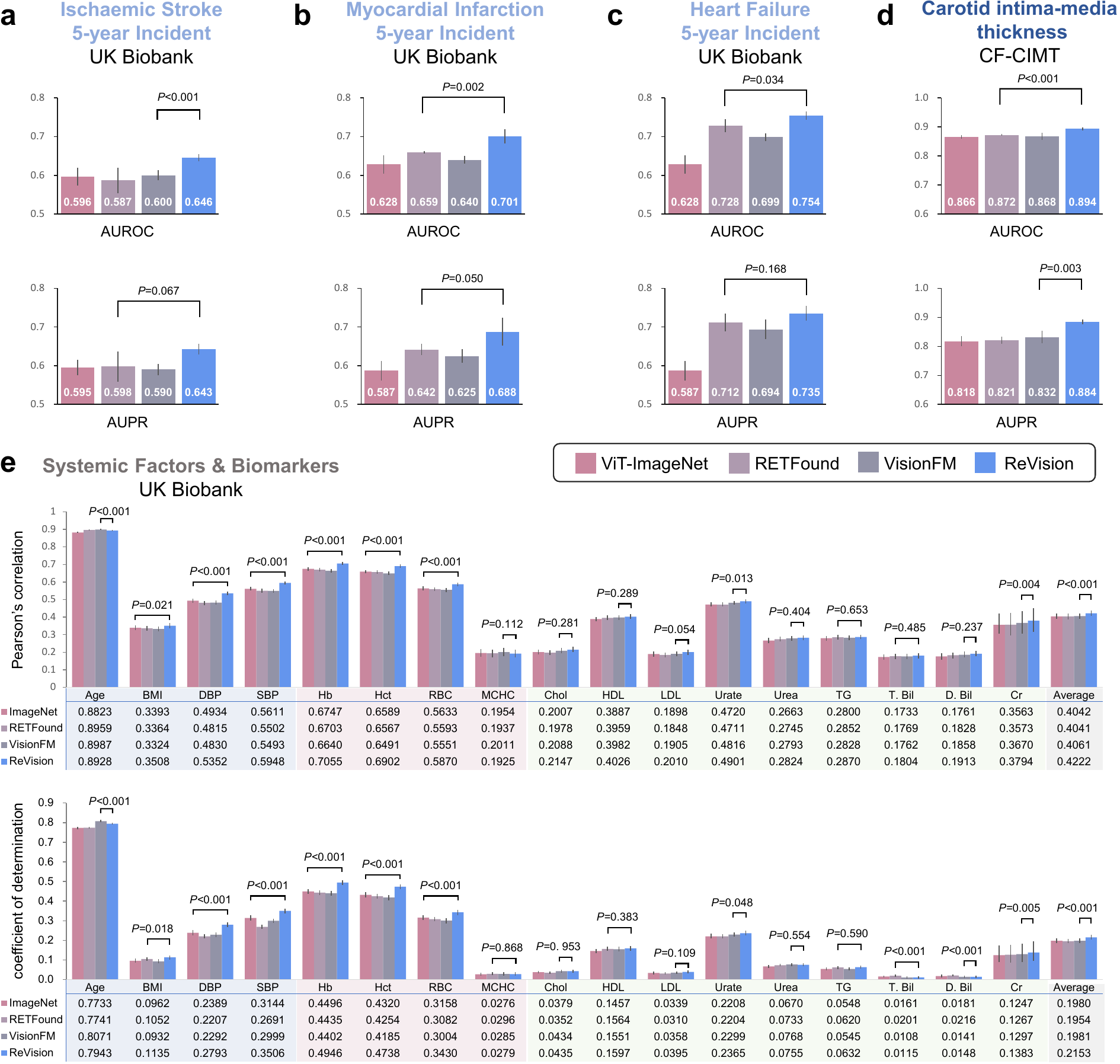}}
\caption{\textbf{Performance on oculomics.} \textbf{a-c,} Results of 5-year incidence prediction of three cardiovascular diseases (CVDs). \textbf{d,} Results of classifying normal carotid intima-media thickness (CIMT) group (< 0.9mm) and thickened CIMT group ($\geq $0.9mm). \textbf{e,} Pearson's correlation and coefficient of determination ($R^2$) of 17 biomarkers. For the UK Biobank, we conducted an external evaluation using data from the Birmingham assessment center while training on data from other centers. In \textbf{a-d}, results were mean across 5 independent experiments, and the error bars show 95\% CI. \textit{P}-value was calculated with the two-sided t-test. In \textbf{e}, 95\% CIs and \textit{P}-values were estimated using the bootstrap method (n=2,000 replicates). BMI, body mass index; DBP, diastolic blood pressure; SBP, systolic blood pressure; Hb, Haemoglobin concentration; Hct, haematocrit percentage; RBC, red blood cell count; MCHC, mean corpuscular haemoglobin concentration; Chol, cholesterol; HDL, HDL cholesterol; LDL, direct LDL cholesterol; TG, triglycerides; T. Bil, total bilirubin; D. Bil, direct bilirubin; Cr, creatinine.
 }
\label{Fig_ex_omics}
\end{figure}

\end{document}

%% file: introduction.tex
\section*{Introduction}

\begin{figure}[htbp]
\setlength{\abovecaptionskip}{0mm}
\centerline{\includegraphics[width=0.9\linewidth]{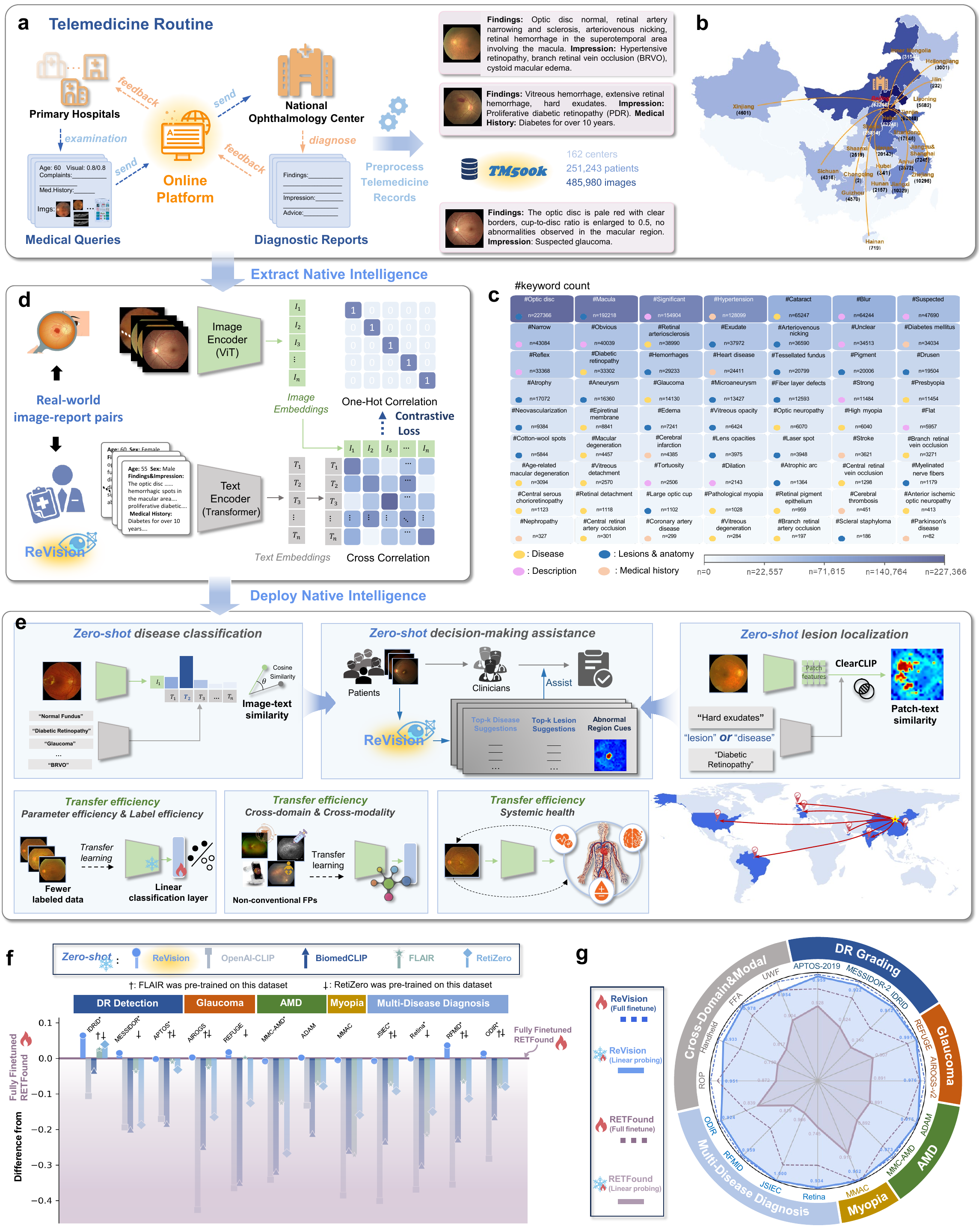}}
\caption{\textbf{Overview of the study.} \textbf{a,} The curation of pre-training dataset from real-world telemedicine practice (TM500k). Through an online consultation platform, primary hospitals submit clinical queries including patient demographics, visual acuity, intraocular pressure, chief complaints, self-reported medical histories, and ophthalmic images. Senior ophthalmologists at a national tertiary ophthalmology center provide comprehensive diagnostic reports encompassing findings, impressions, and clinical advices. The textual reports are extracted, standardized to eye-level granularity, and translated to English via GPT-4o. \textbf{b,} Geographic distribution of the {TM500k}, spanning 162 primary healthcare institutions across 22 provinces in China. \textbf{c,} Visual-language pre-training paradigm for learning native intelligence. ReVision employs contrastive learning to align visual and textual embeddings from authentic clinical image-report pairs while distinguishing unrelated pairs. \textbf{d,} Frequency distribution of clinical terminology in {TM500k} textual reports, illustrating the rich diversity of pathologies encountered in real-world practice. \textbf{e,} ReVision is evaluated across a comprehensive suite of clinical tasks. \textbf{f,} Zero-shot disease detection performance. ReVision achieves remarkable out-of-the-box diagnostic capability without any task-specific labels or optimization. \textbf{g,} Downstream transfer learning performance across 16 diverse ophthalmic tasks. Training only a linear classifier on ReVision's embeddings (linear probing) surpasses fully fine-tuned RETFound \cite{zhou2023foundation}. }
\label{Fig1}
\end{figure}


Ophthalmic foundation models \cite{zhou2023foundation,qiu2023visionfm,silva2025foundation,wang2024retizero,eyefm} (FMs) have demonstrated substantial potential for providing generalizable representations, yet their construction and deployment remain constrained.
From a model construction perspective, existing approaches either pre-train solely on isolated images without clinical context\cite{zhou2023foundation,qiu2023visionfm,sun2025data}, or rely on public research datasets where disease labels are converted into text descriptions through predefined templates\cite{silva2025foundation,keepfit,wang2024retizero}. Both paradigms fail to capture the authentic diagnostic reasoning (from visual appearance to findings and diagnosis) that characterizes real-world medical practice. From a model deployment perspective, these FMs still require substantial annotated data and computational resources for each new application due to the lack of clinical context during pre-training. This gap between model capabilities and deployment requirements represents a primary barrier to translating FMs into clinical practice, particularly in resource-constrained settings.

We reasoned that this gap could be addressed by building clinical native intelligence directly from authentic clinical documentation rather than engineered datasets. Our key insight is that telemedicine programs represent an untapped reservoir for learning image-to-diagnosis reasoning. Such programs offer a unique combination: images acquired from diverse clinical scenarios with varying equipment and patient populations, paired with standardized interpretations from experienced specialists at centralized centers. Such image-report pairs capture authentic image interpretations as they naturally occur in real-world clinical workflows, encoding mutual information between visual appearance, lesion identification, disease impression, and clinical context that image-only and label-based pre-training cannot replicate.

Here we present ReVision, the first retinal FM grounded in large-scale, real-world telemedicine records. We leveraged a collaborative telemedicine program spanning most regions of China (Fig.~\ref{Fig1}a, b).  Over an 8-year period, senior ophthalmologists at a national tertiary center provided remote consultations for primary hospitals, generating comprehensive reports including clinical findings, disease impressions, and management recommendations. After automated preprocessing to ensure data quality, we curated the largest known collection of retinal images with corresponding diagnostic reports, containing 485,980 color fundus photographs (CFPs) spanning a broad range of pathologies  (Fig.~\ref{Fig1}c). Building upon this rich resource, ReVision employs the contrastive language-image pre-training (CLIP\cite{radford2021clip}) to align visual features with textual interpretations while distinguishing unrelated image-text pairs  (Fig.~\ref{Fig1}d).

We evaluated ReVision across 27 benchmarks spanning seven countries to assess deployment efficiency along various dimensions (Fig.~\ref{Fig1}e). First, for zero-shot image understanding, ReVision achieves an AUROC of 0.946 across 12 public benchmarks and 0.952 on 3 real-world multi-disease cohorts without any task-specific training (Fig.~\ref{Fig1}f). The detection performance not only surpasses ophthalmic vision-language FMs \cite{silva2025foundation,wang2024retizero}, but also approaches the fully fine-tuned RETFound model\cite{zhou2023foundation}. Zero-shot pathology localization further provides interpretable visual evidence for predictions. 
When minimal adaptation is feasible, ReVision demonstrates remarkable transfer efficiency. For parameter efficiency, training only a single classification layer outperforms fully fine-tuned competing models on 11 of 12 benchmarks while reducing trainable parameters by 10,000-fold (Fig.~\ref{Fig1}g). For label efficiency, ReVision trained on 10\% of available labels surpasses the best alternative trained on complete datasets. The learned representations also generalize effectively to new clinical sites, imaging domains (hand-held device, premature infants), imaging modalities (ultra-widefield photography, fluorescein angiography), and novel oculomics tasks.
To validate that this deployment efficiency translates to real clinical utility, we conducted a prospective reader study involving 33 ophthalmologists, where ReVision's zero-shot assistance improved diagnostic accuracy from 58.4\% to 73.2\% across all experience levels. Additionally, through detailed analysis of human-AI collaboration behaviors and subjective perceptions from clinicians, we provided new insights into the AI-assisted clinical decision-making workflow. Together, these results demonstrate that large-scale, well-documented clinical archives provide an effective source for building medical AI systems suited to diverse deployment scenarios with limited local resources.

%% file: results.tex
\section*{Results}

\begin{figure}[!t]
\setlength{\abovecaptionskip}{0mm}
\centerline{\includegraphics[width=0.95\linewidth]{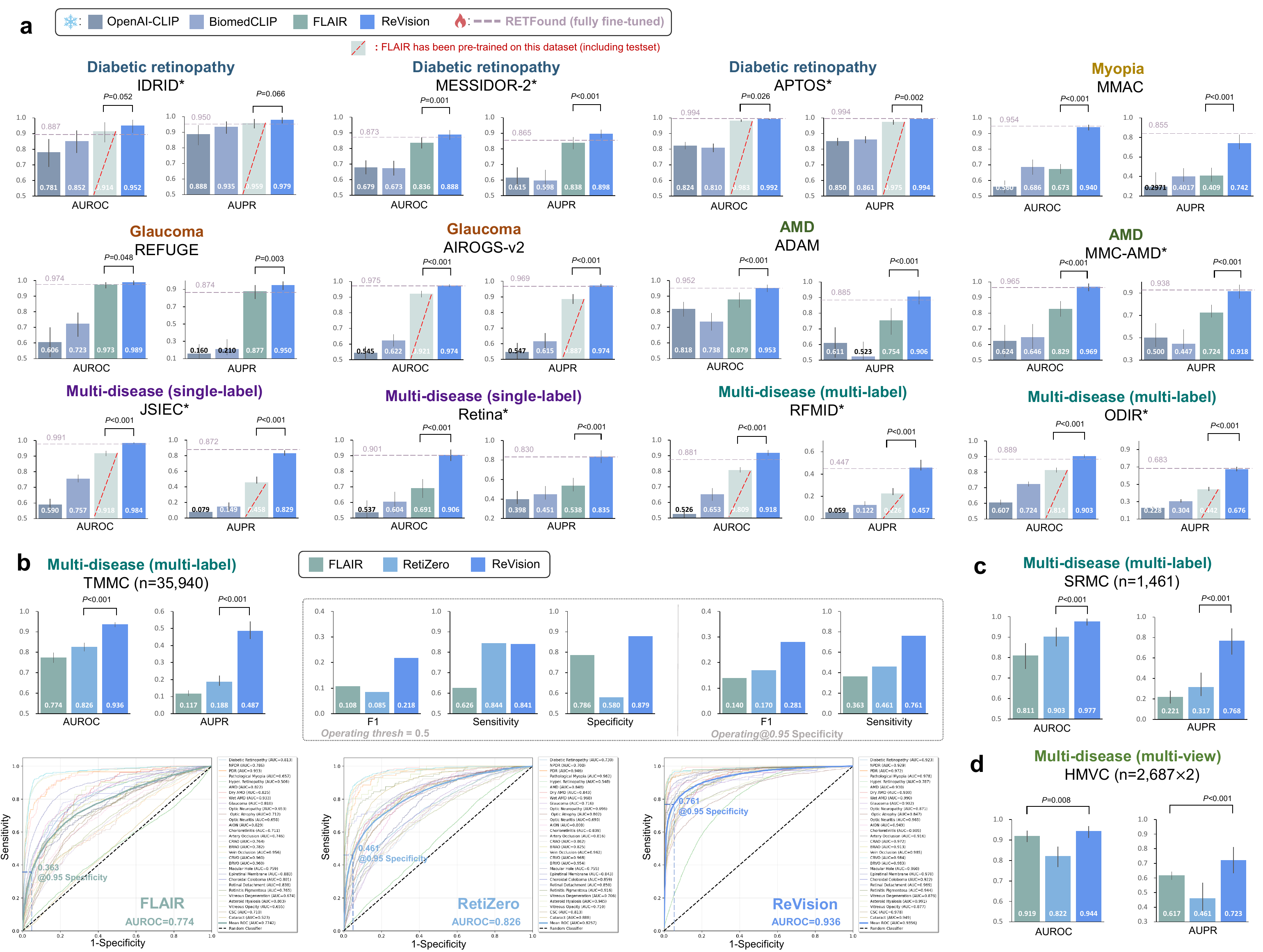}}
\caption{\textbf{Zero-shot disease detection.} \textbf{a,} Performance comparison on public benchmarks. ReVision can perform disease diagnosis without any task-specific optimization, achieving performance comparable to fully fine-tuned RETFound on corresponding downstream tasks. *: To build unambiguous zero-shot disease detection benchmarks, we trimmed the original datasets. For the DR grading datasets, we transformed them into binary DR detection (DR grading is shown in Extended Data Fig. \ref{fig_ex_drgrad}). For MMC-AMD, we merged the "PCV" class (sub-type of wet-AMD) into wet-AMD to avoid ambiguity. For Retina, RFMID, and ODIR, we ignored the undefined "other diseases" class. \textbf{b}, Performance comparison on the Telemedicine Multi-disease Multi-label Cohort (TMMC). \textit{Upper panels}: Class-average (macro) performance metrics showing AUROC, AUPR, accuracy, sensitivity, specificity  (at operating threshold = 0.5), accuracy at 95\% specificity, and sensitivity at 95\% specificity. \textit{Lower panels}: Disease-specific ROC curves for all 30 ophthalmic conditions. ReVision (right) shows consistently superior performance. \textbf{c}, Performance comparison on the Shanghai Ruijin Multi-class Cohort (SRMC). \textbf{d}, Performance comparison on the Handan Multi-view Cohort (HMVC). Non-parametric bootstrapping with 2,000 resamples is employed to derive 95\% CIs and \textit{P}-values. 
}
\label{Fig2}
\end{figure}

\begin{figure}[htbp!]
\setlength{\abovecaptionskip}{0mm}
\centerline{\includegraphics[width=0.9\linewidth]{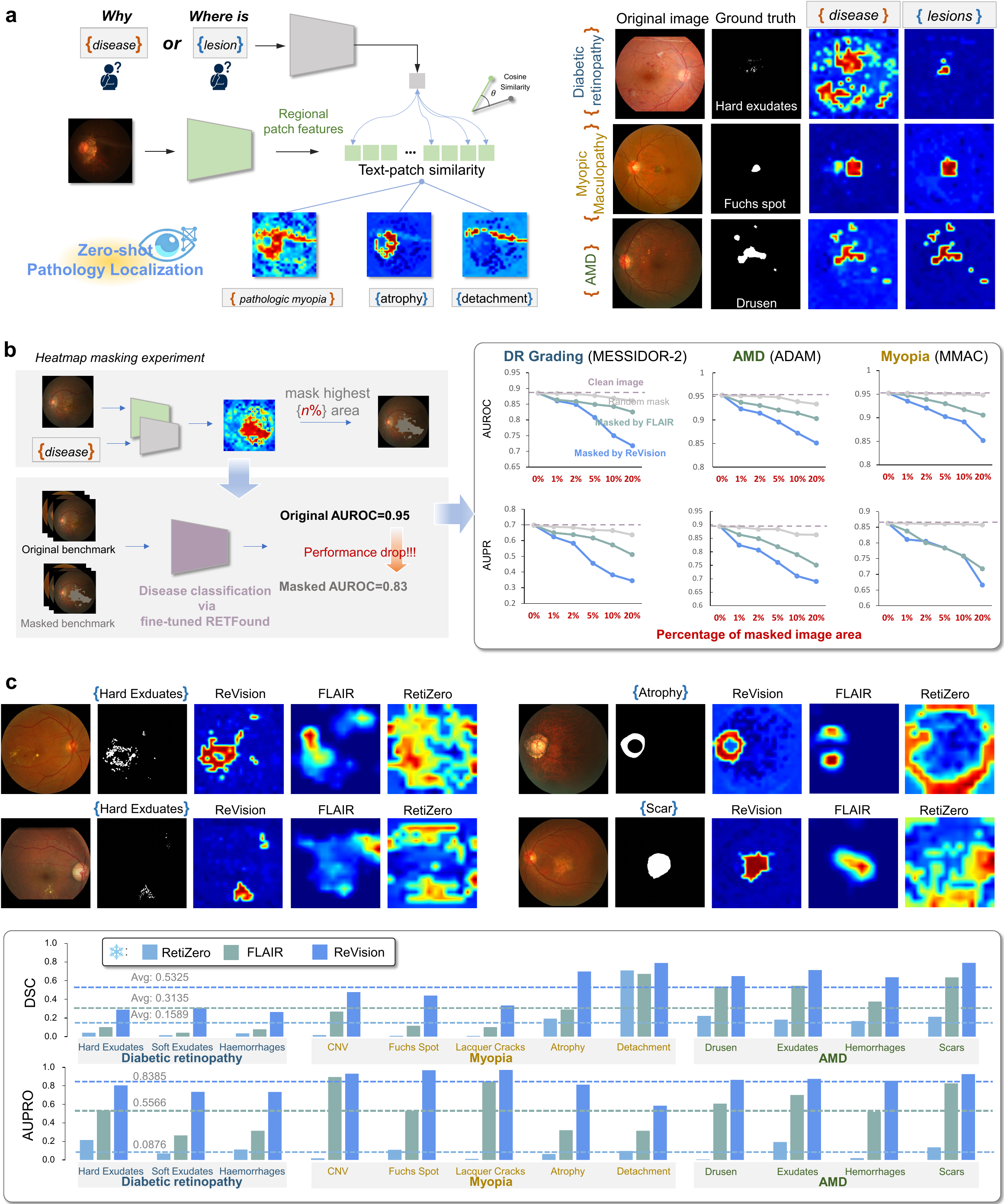}}
\caption{\textbf{Zero-shot pathology localization.}  \textbf{a,} Illustration of zero-shot pathology localization.  When prompted with disease-level descriptions, ReVision localizes pathological evidence indicative of the diagnosis. When prompted with lesion-level descriptions, it identifies the specific abnormalities precisely. \textbf{b,} Explainability demonstration of how our localization maps affect disease diagnosis. On three disease classification datasets, we plot the performance degeneration of fine-tuned RETFound at different region-masking percentages using heatmaps identified by our ReVision or FLAIR. A steeper degeneration implies more informative pathology maps and more interpretative models. \textbf{c,} Qualitative and quantitative assessments of zero-shot lesion localization. Results demonstrate that ReVision substantially outperforms existing retinal vision-language pre-training models.
}
\label{Fig3}
\end{figure}

\subsection*{ReVision delivers zero-shot image understanding}

\textbf{Zero-shot disease detection.}
A hallmark of deployable FM is its zero-shot diagnostic ability\cite{radford2021clip}, recognizing disease categories without training on new labeled examples. This capability ensures robust performance across diverse clinical settings and enables seamless deployment to previously unseen data domains. To evaluate ReVision's zero-shot performance, we leveraged its inherent vision-language alignment: diagnostic categories (\textit{e.g.}, "normal fundus", "diabetic retinopathy") were encoded as text prompts, and classification was performed by computing cosine similarities between image and text embeddings in the unified space.

Across 12 diagnostic benchmarks, ReVision demonstrated exceptional zero-shot performance with an average area under the receiver operating characteristic curve (AUROC) of 0.946 (Fig.~\ref{Fig2}a). ReVision surpassed general-purpose and biomedical vision-language models (VLMs) substantially in zero-shot settings, achieving 47.5\% and 33.8\% higher average AUROCs versus OpenAI-CLIP\cite{radford2021clip} and BiomedCLIP\cite{zhang2023biomedclip} across 12 benchmarks. When compared to specialized ophthalmic vision-language models, ReVision maintained consistent superiority\cite{silva2025foundation,wang2024retizero}. Although FLAIR\cite{silva2025foundation} demonstrated competitive performance on certain datasets, its pre-training exposure to 6 of our 12 evaluation benchmarks potentially compromises zero-shot validity; nevertheless, ReVision outperformed FLAIR by 10.5\% AUROC across all benchmarks and 15.1\% on the 6 datasets unseen during FLAIR's training. Given RetiZero's\cite{wang2024retizero} extensive pre-training on 29 public datasets (including 9 of our 12 benchmarks), we restricted comparisons to the 4 datasets absent from its training data, where ReVision demonstrated 21.4\% higher AUROC (Extended Data Fig.~\ref{Fig_ex_4datazs}). Notably, on 11 of 12 benchmarks, zero-shot ReVision matched or exceeded the performance of RETFound\cite{zhou2023foundation} fully fine-tuned on target datasets, demonstrating that ReVision has learned transferable disease representations that enable immediate deployment without local data annotation

 Beyond public research benchmarks, we assessed ReVision under authentic clinical conditions using three independent cohorts. The Telemedicine Multi-disease Multi-label Cohort (TMMC) was held out from pretraining and comprises multi-label 30 disease categories. ReVision achieved an AUROC of 0.938 and sensitivity of 0.761 at 95\% specificity, substantially outperforming FLAIR (AUROC: 0.774, sensitivity: 0.346) and RetiZero (AUROC: 0.826, sensitivity: 0.461) (Fig.~\ref{Fig2}b). Similar advantages were observed on the Shanghai Ruijin Multi-class Cohort (SRMC; AUROC: 0.977) representing routine outpatient care (Fig.~\ref{Fig2}c), and the Handan Multi-view Cohort (HMVC; AUROC: 0.944) representing opportunistic screening with multi-view imaging (Fig.~\ref{Fig2}d).

\textbf{Zero-shot pathology localization.}
Interpretability is essential for clinical deployment. Using attention-based class activation mapping (ClearCLIP\cite{lan2024clearclip}), we investigated whether ReVision's predictions are grounded in clinically relevant image regions. When prompted with disease names, ReVision localized pathological regions accurately; when prompted with specific lesions (e.g., "hard exudates," "drusen"), it generated precise activation maps delineating corresponding structures (Fig.~\ref{Fig3}a and Extended Data Fig. \ref{Fig_ex_zs_loc}).

We validated these localizations through two experiments. First, a masking study assessed whether disease-level activations capture diagnostically critical regions. On three classification benchmarks, we progressively masked the top-activated pixels and evaluated a task-specific classifier\cite{zhou2023foundation} on the masked images. Masking 10\% of ReVision-identified regions caused AUROC drops of 6.4–15.3\%, significantly exceeding the impact of masking FLAIR-identified or random regions (Fig.~\ref{Fig3}b). Second, we evaluated lesion-level localization across 12 pathological features in diabetic retinopathy, AMD, and pathologic myopia. Without any segmentation training, ReVision achieved a mean Dice Similarity Coefficient (DSC) of 0.533 and Per-Region Overlap score (PRO) of 0.839, substantially outperforming FLAIR (DSC: 0.314, PRO: 0.557) and RetiZero (DSC: 0.159, PRO: 0.088) (Fig.~\ref{Fig3}c). These results demonstrate that vision-language alignment with detailed clinical reports enables ReVision to provide interpretable, clinically meaningful localizations.

\subsection*{ReVision enables efficient transfer learning}

\begin{figure}[htbp!]
\setlength{\abovecaptionskip}{0mm}
\centerline{\includegraphics[width=0.95\linewidth]{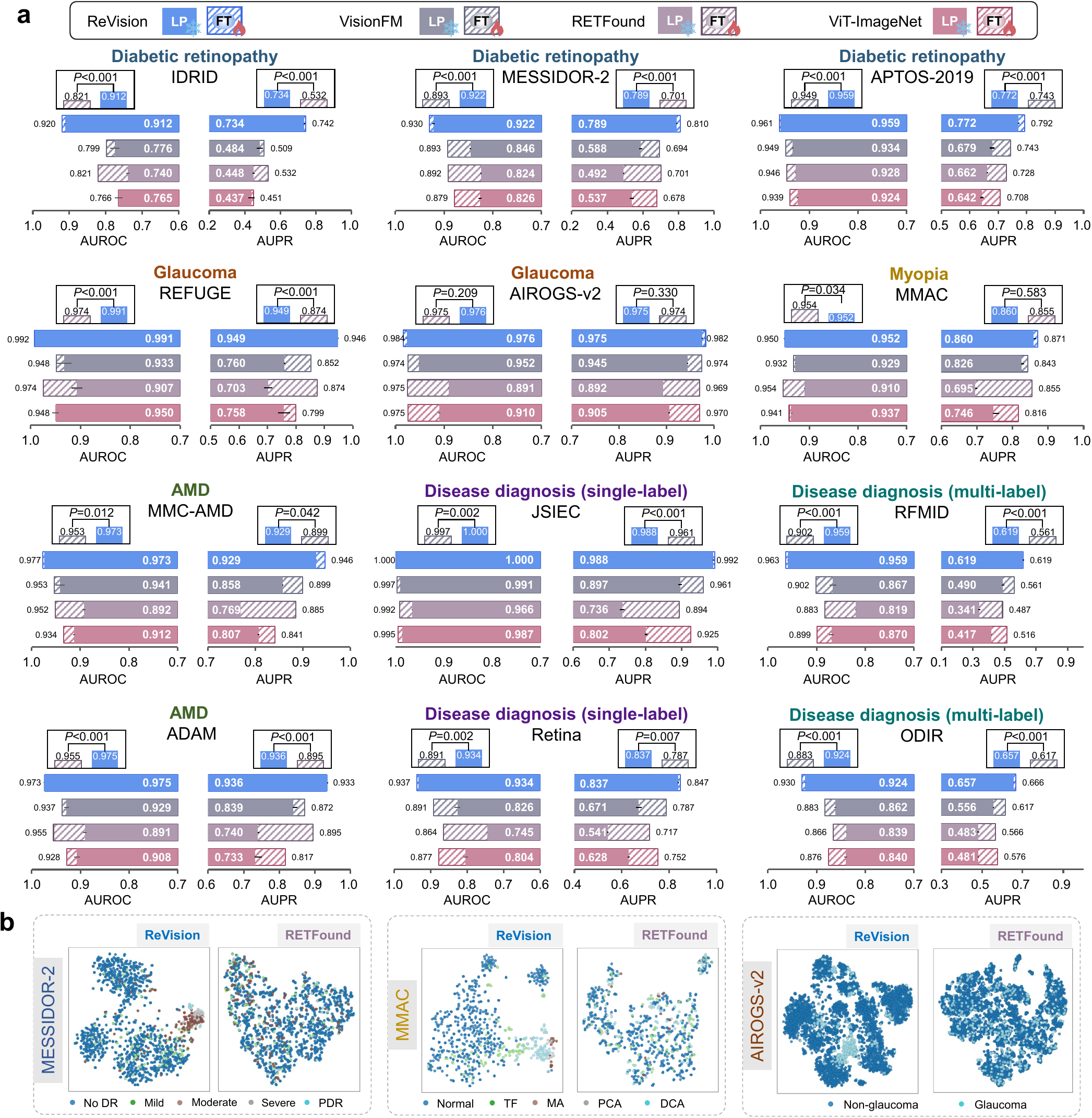}}
\caption{\textbf{Transfer learning via linear probing.} \textbf{a,} Ocular disease diagnosis results measured by AUROC and AUPR. Solid bars and outlined bars indicate linear probing (LP) and full fine-tuning (FT), respectively. The top boxed annotations present the statistical significance (P-values) of ReVision's LP performance compared to the best FT performance from the comparison models. Reported values represent the mean of five independent runs, with error bars denoting the 95\% CI. \textit{P}-values are computed using the two-sided t-test. \textbf{b,} t-SNE visualization of image embeddings from pre-trained encoders without further fine-tuning. Each point represents an image, and different colors denote different classes (see Extended Data Fig. \ref{Fig_ex_tnse} for more datasets). The more distinct clustering patterns demonstrate that ReVision learns more discriminative representations. PDR, proliferative DR; AMD, age-related macular degeneration; TF, tessellated fundus; MA, macular atrophy; PCA, patchy chorioretinal atrophy; DCA, diffuse chorioretinal atrophy.}
\label{Fig_ftlp}
\end{figure}

\textbf{Parameter efficiency via linear probing.}
When labeled data is available for specific tasks, fine-tuning (FT) all model parameters of an FM typically achieves optimal performance but requires substantial computational resources—a barrier for resource-constrained practitioners without access to advanced computing infrastructure. In contrast, linear probing (LP) offers significant computational advantages by training only a classification layer while freezing the FM's weights. We compared both approaches against state-of-the-art models (RETFound\cite{zhou2023foundation}, VisionFM\cite{qiu2023visionfm}, ViT-ImageNet\cite{vit2021}) across 12 disease classification benchmarks.
Most notably, LP ReVision ($\#classes\times 1024$ trainable parameters) achieved superior performance on 11 of 12 benchmarks compared to any FT competitor  (Fig.~\ref{Fig_ftlp}a), with an average of 3.6\% AUROC and 10.3\% AUPR improvements over FT RETFound ($\sim 3.3 \times 10^8$ trainable parameters). This efficiency extended to segmentation tasks, where ReVision achieved 0.651 average DSC across 12 lesion segmentation tasks using a segmentation head on the top of the frozen backbone (Extended Data Fig.~\ref{Fig_ex_seg}). To better understand why LP ReVision achieves superior performance, we visualized the feature embeddings of ReVision and RETFound using t-distributed Stochastic Neighbor Embedding (t-SNE) (Fig.~\ref{Fig_ftlp}b and Extended Data Fig. \ref{Fig_ex_tnse}). The visualization revealed that ReVision's pretrained embeddings form notably more distinct disease clusters than RETFound, explaining why a simple linear classifier suffices.

\begin{figure}[!t]
\setlength{\abovecaptionskip}{0mm}
\centerline{\includegraphics[width=0.95\linewidth]{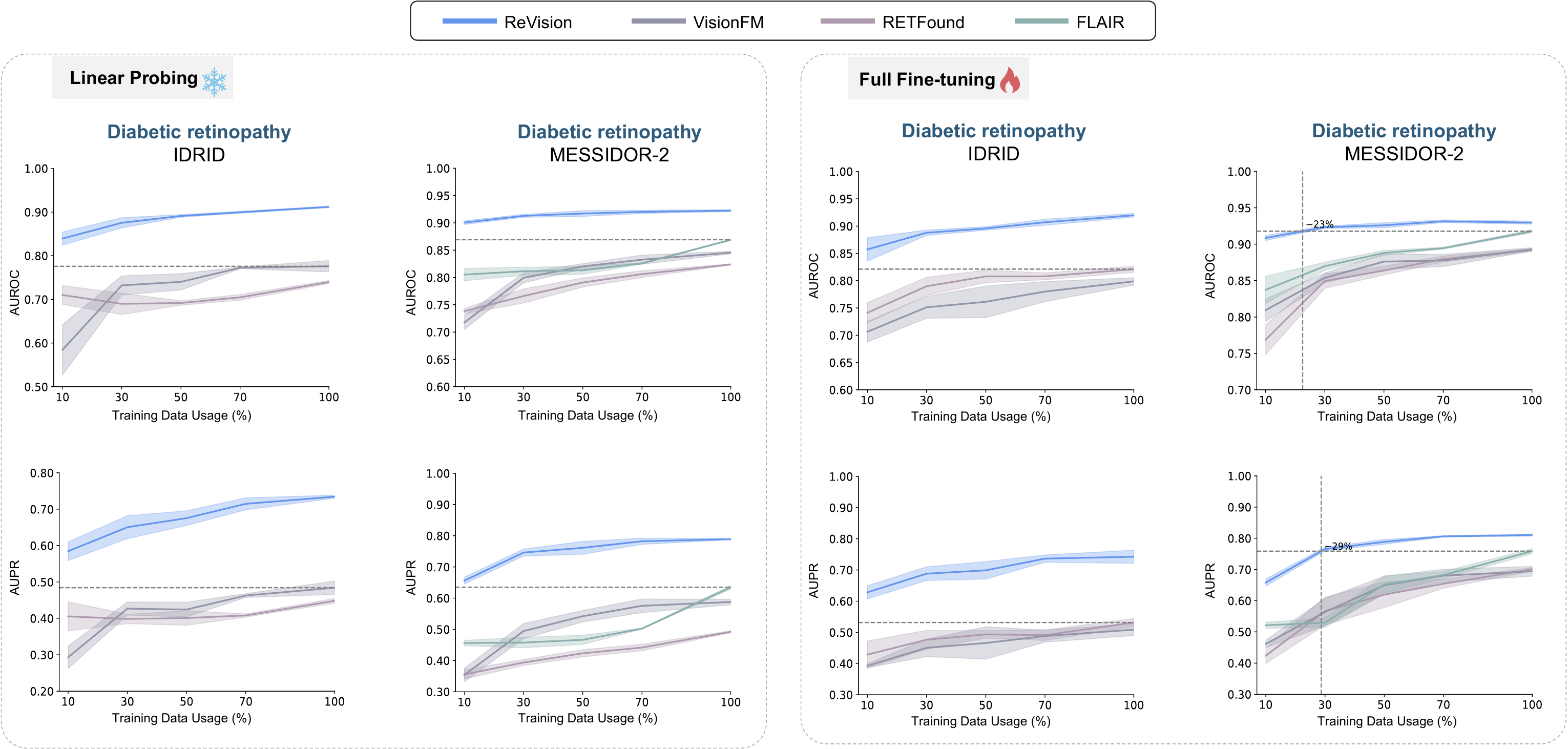}}
\caption{\textbf{Label efficiency during downstream training}. Center lines represent mean values across 5 runs, with colored bands indicating 95\% confidence intervals. The dashed reference lines highlight that ReVision achieves comparable or superior performance to fully-trained comparison models using only a small fraction of the training data. The pre-training data of FLAIR has included IDRID. 
}
\label{Fig_label_eff}
\end{figure}

\textbf{Label efficiency with limited annotations.}
Clinical data annotation is expensive and time-consuming, making label-efficient learning essential for practical deployment. We evaluated ReVision's efficiency using reduced training data fractions (10\%, 30\%, 50\%, 70\%) across DR grading benchmarks. With only 10\% of training data, LP ReVision achieved AUROCs of 0.839 (95\% CI: 0.824-0.854) and 0.901 (95\% CI: 0.897-0.904) on IDRID and MESSIDOR-2, outperforming LP RETFound trained on full data by 9.9\% and 7.7\% (\textit{P} < 0.001) under the LP protocol (Fig.~\ref{Fig_label_eff}). Remarkably, this 10\%-data LP ReVision even surpassed FT RETFound trained on 100\% data, demonstrating that ReVision's superior embeddings can compensate for both limited training data and minimal parameter adaptation. ReVision also exhibited significantly faster convergence, requiring only 5\% of the training epochs needed by RETFound to achieve comparable performance (Extended Data Fig.~\ref{Fig_ex_converge}).

\textbf{Generalization efficiency to new sites.}
After being fine-tuned to a new task, robust AI systems must maintain performance when deployed across different healthcare institutions with varying patient populations, imaging protocols, and equipment configurations. We conducted cross-dataset evaluations (external evaluation) across IDRID\cite{IDRID}, MESSIDOR-2\cite{MESSIDOR2}, and APTOS-2019\cite{APTOS} for DR grading, and between REFUGE\cite{REFUGE} and AIROGS\cite{AIROGS} for glaucoma detection. When supervised on IDRID and tested on MESSIDOR-2, ReVision achieved an AUROC of 0.895 (95\% CI: 0.893-0.897), significantly outperforming RETFound (0.831) and VisionFM (0.763), with consistent advantages across all evaluations (Extended Data Fig.~\ref{fig_ex_external}). Notably, LP ReVision performed comparably to FT Revision under domain shifts (mean AUROC difference = 0.6\%), suggesting that ReVision's frozen embeddings already encode generalizable features, while task-specific fine-tuning may compromise cross-domain transferability.

\begin{figure}[!t]
\setlength{\abovecaptionskip}{0mm}
\centerline{\includegraphics[width=0.99\linewidth]{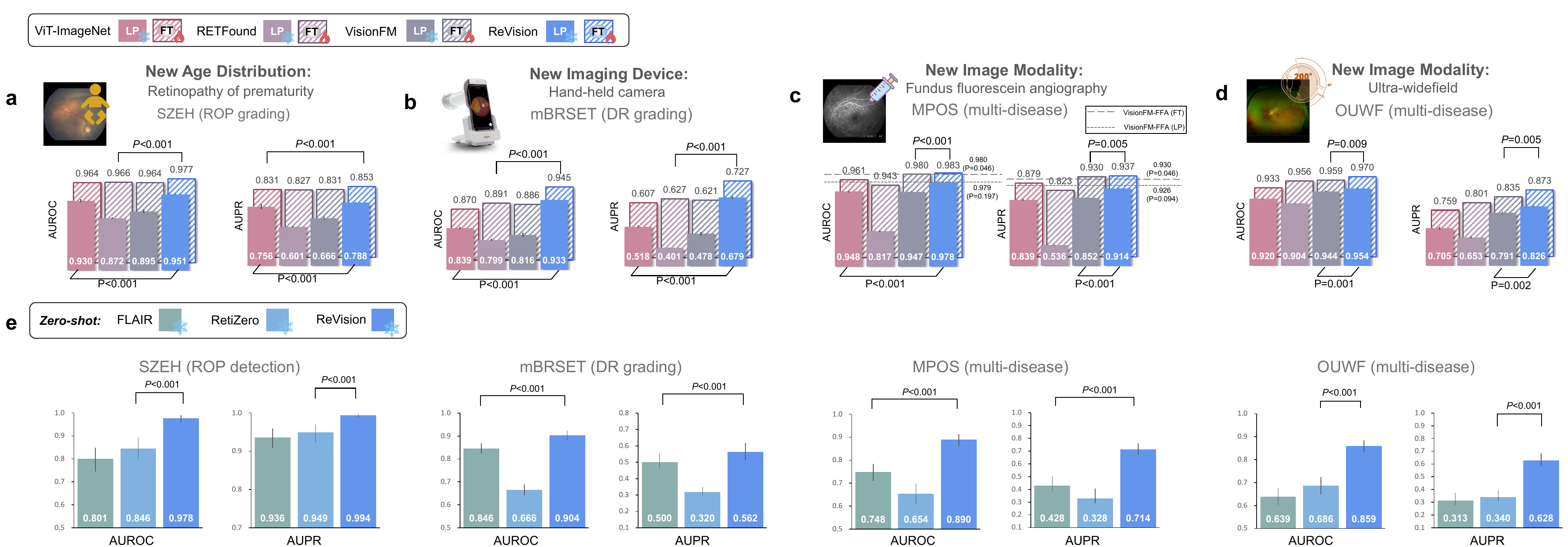}}
\caption{\textbf{Transfer efficiency for cross-domain and cross-modality learning.} \textbf{a-d,} Supervised downstream transfer learning on near-CFP domains and unseen ophthalmic imaging modalities. \textbf{a,} Performance on the SZEH dataset\cite{ROP} for retinopathy of prematurity (ROP) grading in premature infants. \textbf{b,} Performance on the mBRSET dataset\cite{mBRSET} for DR grading. CFPs were captured using the Phelcom Eyer (Phelcom Technologies, São Carlos, Brazil), a portable, handheld, smartphone-based retinal fundus camera. \textbf{c,} Multi-disease diagnosis on fundus fluorescein angiography (FFA) fundus using MPOS\cite{MPOS} dataset that contains 5 categories. \textbf{d,} Multi-disease classification on ultra-widefield (UWF) fundus images using OUWF\cite{OUWF} dataset that contains 7 categories. VisionFM provides a version pre-trained on FFA, which we also included for comparison. For a-d, solid bars represent linear probing (LP) and outlined bars represent full fine-tuning (FT). Reported values represent the mean of five independent runs. \textit{P}-values are computed using the two-sided t-test. \textbf{e,} Zero-shot performance across all four cross-domain/modal benchmarks, demonstrating ReVision's ability to generalize without any parameter updates or task-specific training. Non-parametric bootstrapping with 2,000 resamples is employed to derive 95\% CIs and \textit{P}-values.}
\label{Fig_modality}
\end{figure}

\textbf{Transfer efficiency beyond standard CFPs.}
We hypothesized that ReVision's exposure to diverse fundus images enables it to capture essential anatomical features shared across different \textit{en face} retinal imaging modalities, transcending standard CFP characteristics. Evaluation across four challenging cross-domain benchmarks confirmed this hypothesis. For retinopathy of prematurity (ROP) in premature infants  (SZEH\cite{ROP}), LP ReVision achieved an AUROC of 0.951 (Fig.~\ref{Fig_modality}a), significantly surpassing other FMs. On portable fundus cameras (mBRSET\cite{mBRSET}), ReVision maintained an AUROC of 0.933 for DR grading (Fig.~\ref{Fig_modality}b). For fundamentally distinct modalities—fundus fluorescein angiography (FFA) capturing vascular dynamics in grayscale and ultra-widefield imaging (UWF) extending the field-of-view to 200°—LP ReVision achieved AUROCs of 0.978 (MPOS\cite{MPOS}) and 0.954 (OUWF\cite{OUWF}), respectively (Fig.~\ref{Fig_modality}c-d). Remarkably, ReVision also demonstrated meaningful zero-shot performance across all four benchmarks (Fig.~\ref{Fig_modality}e), including 0.978 AUROC (95\% CI: 0.960-0.991) for ROP detection without any infant CFP training, substantially outperforming FLAIR, and RetiZero. This efficiency across diverse imaging conditions demonstrates that ReVision has learned fundamental retinal pathology principles that transcend specific protocols.

\begin{figure}[!t]
\setlength{\abovecaptionskip}{0mm}
\centerline{\includegraphics[width=0.99\linewidth]{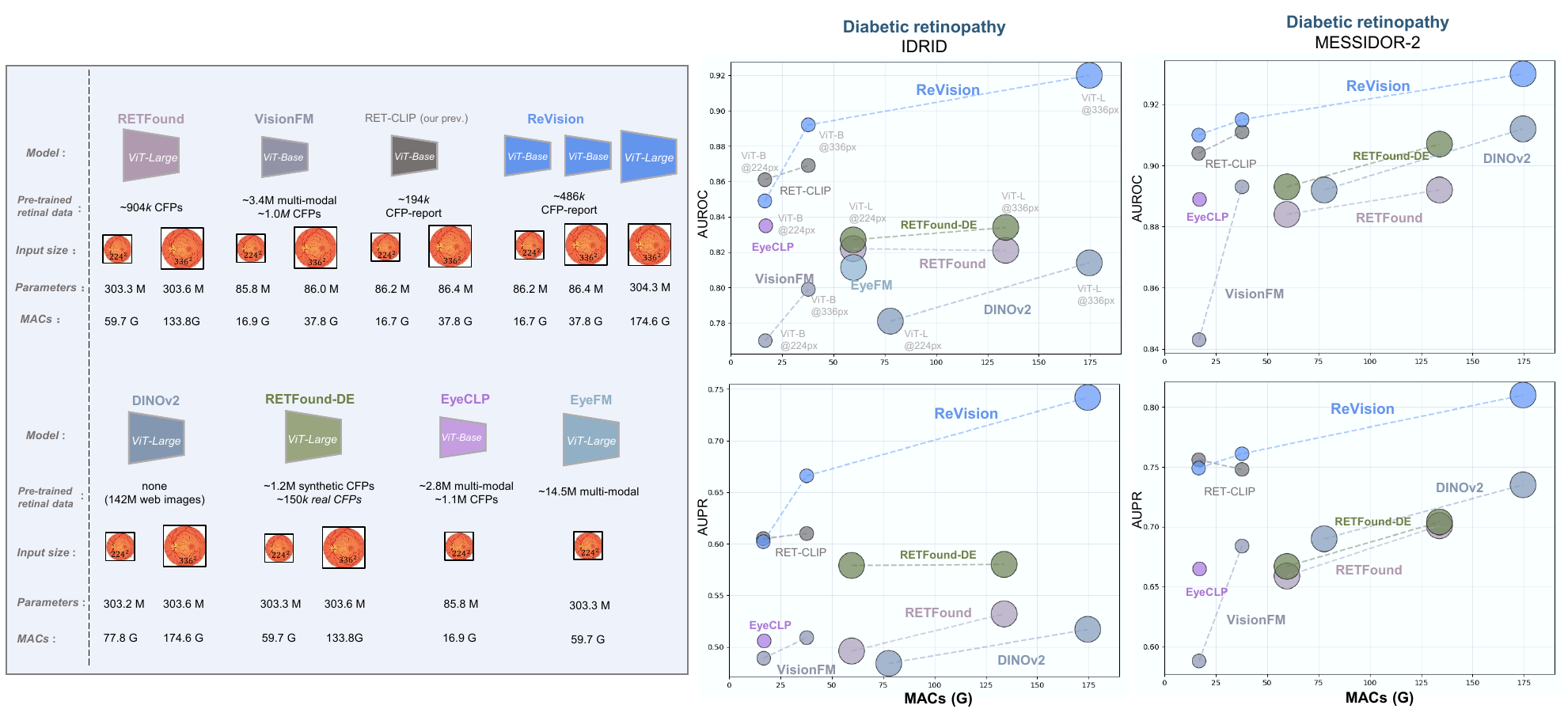}}
\caption{\textbf{Computational efficiency across different model size and image resolution}. We compared ReVision with RETFound\cite{zhou2023foundation}, VisionFM\cite{qiu2023visionfm}, DINOv2\cite{oquab2024dinov2}, RETFound-DE\cite{sun2025data}, EyeCLIP\cite{shi2024eyeclip}, EyeFM\cite{eyefm}, and our previous RET-CLIP\cite{retclip}, across different architectures (ViT-Base and ViT-Large) and input resolutions (224² and 336² pixels). RET-CLIP\cite{retclip} is our preliminary proof-of-concept model trained on an uncleaned subset of Chinese telemedicine data. DINOv2\cite{oquab2024dinov2} is the current state-of-the-art general-purpose vision foundation model pre-trained on a meticulously curated dataset of 142 million web images. RETFound-DE\cite{sun2025data} is a retinal foundation model pre-trained on 1 million synthetic retinal images and 150 thousand real retinal images. EyeCLIP\cite{shi2024eyeclip} is pre-trained on 2.7 million multi-modality ophthalmic images, in which 11,180 angiography images contain textual reports. EyeFM\cite{shi2024eyeclip} is pre-trained on 14.5 million multi-modality ophthalmic images.
}
\label{Fig_scale}
\end{figure}

\textbf{Computational efficiency through model scaling.}
Healthcare institutions operate under varying computational constraints, necessitating flexible model architectures that balance diagnostic accuracy with resource limitations. We pre-trained multiple variants with different architectures (ViT-Large, ViT-Base) and input resolutions (336$\times$336, 224$\times$224 pixels), then assessed performance on IDRID and MESSIDOR-2 using full fine-tuning.
Scaling experiments revealed consistent performance improvements with increased model capacity and resolution (Fig.~\ref{Fig_scale}). Remarkably, even the most efficient ViT-Base@224px variant—requiring only 10\% of computational operations and 28\% of overall parameters—consistently outperformed RETFound (ViT-Large@336px). We further benchmarked ReVision against diverse FMs including DINOv2\cite{oquab2024dinov2}, RETFound-DE\cite{sun2025data}, EyeCLIP\cite{shi2024eyeclip}, EyeFM\cite{eyefm}, and our previous RET-CLIP\cite{retclip}. ReVision maintained significant advantages across all architectures, underscoring the value of large-scale authentic image-report pairs over synthetic data or vision-only pretraining. This flexibility enables practitioners to select optimal accuracy-efficiency trade-offs for their specific resource constraints.

\textbf{Transfer efficiency for systemic health prediction.}
Beyond ocular pathologies, we explored whether representations learned from ophthalmic consultations transfer to oculomics\cite{qian2024cardiovascular,diaz2022predicting,poplin2018prediction,mitani2020detection}. We evaluated ReVision's ability to predict 5-year incident cardiovascular diseases (CVDs) using the UK Biobank cohort\cite{ukbb}, assessing three major outcomes: ischaemic stroke, myocardial infarction, and heart failure. ReVision achieved AUROCs of 0.646, 0.701, and 0.754 for these three conditions, respectively, demonstrating modest improvements over RETFound (Extended Data Fig.~\ref{Fig_ex_omics}a-c).
For carotid intima-media thickness (CIMT) prediction using the CF-CIMT dataset \cite{cimt}, ReVision achieved an AUROC of 0.894, demonstrating potential for non-invasive vascular health assessment (Extended Data Fig.~\ref{Fig_ex_omics}d). Analysis of 17 systemic biomarkers revealed strong correlations with age, blood pressure, and hematological markers (Pearson's correlation >0.5), though correlations with metabolic indicators remained limited (Extended Data Fig.~\ref{Fig_ex_omics}e).
These results demonstrate that representations learned by ReVision can transfer to systemic health prediction tasks, though performance varies with task complexity and available training data. This transfer capability suggests potential applications in screening settings where retinal imaging could provide health insights beyond ocular assessment.

\begin{figure}[!t]
\setlength{\abovecaptionskip}{0mm}
\centerline{\includegraphics[width=0.95\linewidth]{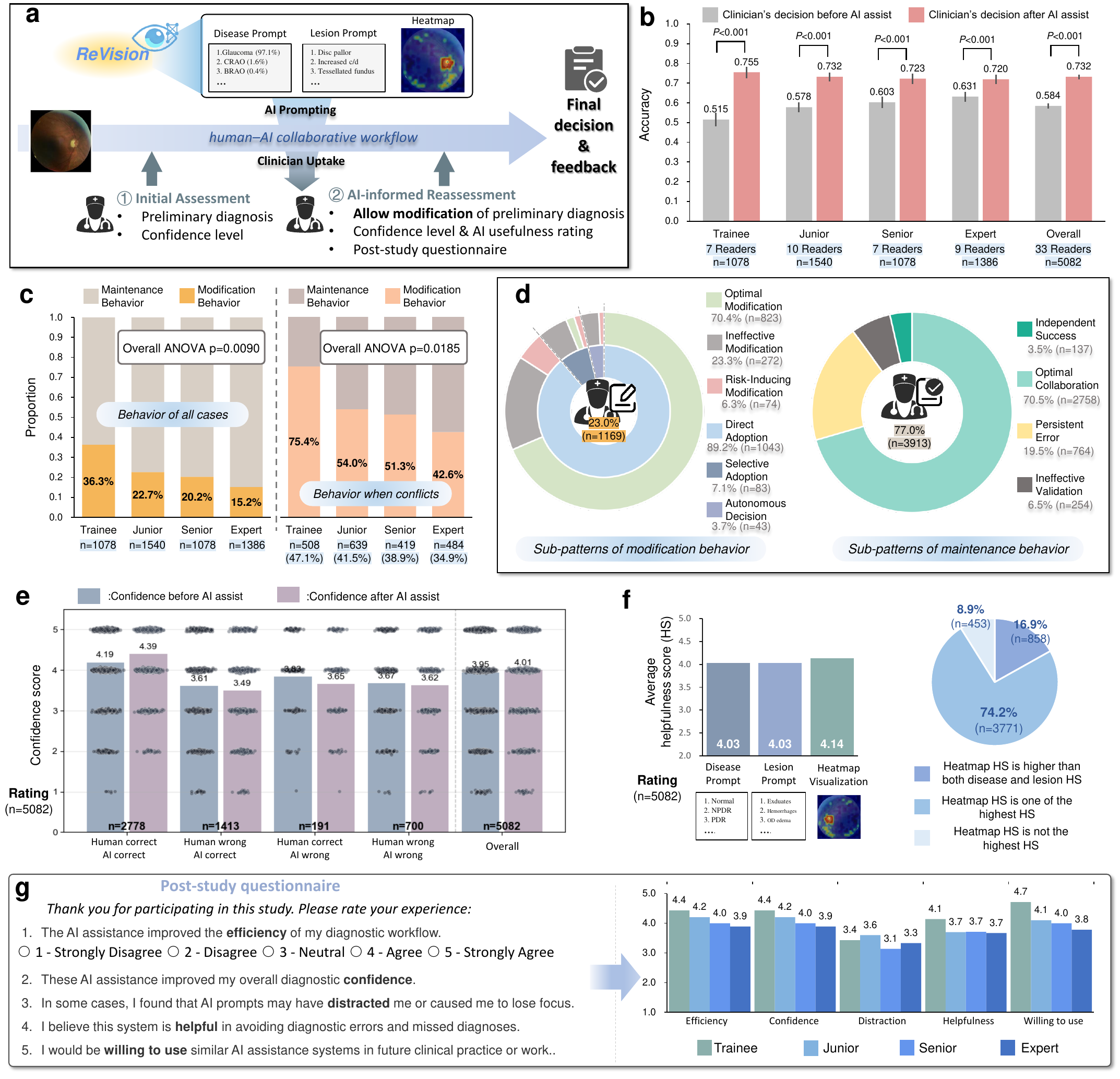}}
\caption{\textbf{Statistical analysis of diagnostic performance, decision behaviors, and subjective perceptions within the human–AI collaborative workflow.} \textbf{a,} Illustration of the human–AI
collaborative workflow. \textbf{b,} Diagnostic accuracy of participants before and after AI assistance within the workflow. Incorporating ReVision into the decision-making process yielded consistent and significant improvements across participants of different experience levels. \textit{P}-values were calculated using McNemar's test. \textbf{c,} Participants' modification behavior within the workflow. Across all cases (left), more experienced participants exhibited fewer modifications, with significant differences observed among the four groups. In cases where ReVision's suggestions conflicted with participants' initial decisions (right), less experienced participants demonstrated higher reliance on ReVision, whereas senior participants tended to trust their own. \textbf{d,} Behavioral pattern analysis of the human–AI collaboration. Among all modification behaviors, participants' decisions were categorized as direct adoption of ReVision's Top-1 suggestion, selective adoption of Top-2–5, or independent re-diagnosis disregarding all Top-5 suggestions, which lead to three distinct outcomes: optimal revision (successfully revising an incorrect preliminary diagnosis), ineffective correction (modification without achieving a correct diagnosis), or risk-inducing revision (overriding a correct preliminary diagnosis with an incorrect one). Among maintenance behaviors, four outcomes were identified: independent success (participants resisted incorrect AI suggestions), optimal collaboration (AI validated a correct preliminary decision), ineffective validation (AI reinforced an incorrect preliminary decision), and persistent error (participants rejected a potentially corrective AI suggestion). \textbf{e,} Changes in participants' diagnostic confidence before and after AI assistance, reflecting subjective perceptions within the workflow. \textbf{f,} Participants' evaluations of the three types of AI-provided information. \textbf{g,} Results of the post-study questionnaire, reflecting participants' subjective acceptance of AI assistance.}
\label{fig_main_rs}
\end{figure}

\subsection*{Zero-shot ReVision improves clinical decision-making}

To validate that ReVision's deployment efficiency translates to real clinical benefit, we conducted a prospective reader study involving 33 ophthalmologists with varying experience levels from multiple institutions. Each participant evaluated 154 CFPs across 21 distinct ocular conditions, yielding 5,082 diagnostic readings. Importantly, ReVision was used in its zero-shot form without any task-specific training, directly testing whether a deployment-efficient model provides meaningful assistance in actual diagnostic practice. The workflow simulated realistic clinical practice (Fig.~\ref{fig_main_rs}a): clinicians first provided independent diagnoses with confidence assessments, then received assistance from ReVision's zero-shot prediction (Top-5 disease predictions, Top-5 lesion identifications, and pathology localization heatmaps) before making final decisions.

Our evaluation encompassed three levels: diagnostic accuracy (outcomes), participants' decision-making behaviors (processes), and participants' subjective perceptions of the workflow.
ReVision-assisted intervention consistently improved diagnostic accuracy across all experience levels (\textit{P} < 0.001), increasing overall accuracy from 58.4\% (95\% CI: 57.6\%–60.4\%) to 73.2\% (95\% CI: 71.9\%-74.4\%) (Fig.~\ref{fig_main_rs}b). Trainees benefited most substantially, with accuracy increasing from 51.5\% to 75.5\%, while improvements progressively diminished for junior physicians (26.6\%), senior physicians (19.9\%), and expert physicians (14.1\%). All groups converged to similar final performance levels (72-76\%), suggesting that AI assistance effectively democratizes diagnostic capability by elevating less experienced clinicians toward expert-level accuracy.

Analysis of decision-making behaviors revealed experience-dependent patterns (Fig.~\ref{fig_main_rs}c,d). Less experienced participants exhibited greater reliance on AI guidance, with trainees modifying 75.4\% of preliminary decisions when disagreeing with AI's top prediction, compared to 42.6\% for experts (Fig.~\ref{fig_main_rs}c). 
Among all modifications (Fig.\ref{fig_main_rs}d, left panel), 70.4\% successfully corrected errors (optimal revision), while 6.3\% overturned correct preliminary diagnoses (risk-inducing revision). A further 23.3\% did not achieve a correct diagnosis but nonetheless triggered clinical attention and re-evaluation (ineffective revision), suggesting practical value beyond accuracy gains. 
Among maintenance behaviors (Fig.\ref{fig_main_rs}d, right panel), 70.5\% yielded optimal collaboration, with correct AI guidance reinforcing participants' initial decisions. However, 19.5\% of maintenance behaviors reflected persistent errors, where mistrust of AI suggestions led to missed opportunities for correction. Importantly, 3.5\% of cases showed independent success where participants appropriately resisted incorrect AI suggestions, demonstrating that physicians maintained critical evaluation rather than blindly following AI outputs.

For participants' subjective perceptions, we evaluated three aspects: diagnostic confidence (Fig.\ref{fig_main_rs}e), ratings of AI-assisted information (Fig.\ref{fig_main_rs}f), and responses to post-study questions (Fig.\ref{fig_main_rs}g). Participants' confidence levels exhibited sophisticated patterns: increasing when AI aligned with their assessments, and decreasing when opinions diverged, serving as a valuable "disagreement as red flag" mechanism for quality control. Among AI-provided information, pathology visualization received the highest utility ratings (4.14/5), followed by lesion identification (4.03/5) and disease predictions (4.03/5), highlighting participants' preference for interpretable outputs. Post-study assessment revealed positive attitudes across all experience levels, with trainee participants providing the most enthusiastic endorsements (4.7/5 willingness to use) while more senior participants approached AI tools more cautiously, primarily viewing them as reference aids.

%% file: discussion.tex
\section*{Discussion}
Existing ophthalmic FMs have long pursued a paradox: achieving clinical deployment readiness through training on curated research datasets that lack authentic clinical context. Vision-only approaches\cite{zhou2023foundation,qiu2023visionfm} learn visual features in a self-supervised manner. While these methods successfully capture visual features, they lack explicit diagnostic knowledge, limiting their ability to provide representations that align with clinical practice. To incorporate textual diagnostic knowledge, vision-language approaches\cite{wang2024retizero,silva2025foundation,keepfit} attempt to convert disease labels from public datasets into text descriptions through predefined templates, while some methods\cite{shi2024eyeclip} further learn from aligning representations across imaging modalities of the same eye. However, such methods face inherent limitations: label ambiguity from single-task annotations that ignore comorbidities, clinical misalignment from synthetic descriptions that enumerate lesions not actually present, and information incompleteness from template-based text that cannot capture diagnostic reasoning. A critical gap across all approaches is the absence of large-scale, authentic clinical reports. However, unlike radiology, where examination reports are routinely written by radiologists\cite{johnson2023mimic,bustos2020padchest}, many ophthalmic images captured during clinical examinations are directly interpreted by the attending ophthalmologist at the point of care, and thus exist without documented imaging reports. Our work addresses this gap by leveraging telemedicine archives where remote consultations naturally generate image-report pairs encoding diagnostic reasoning.  
In our exploratory works\cite{retclip,yang2024vilref}, we conducted preliminary investigations leveraging a subset of original Chinese telemedicine data and small models to validate this concept. Building upon these experiences and insights, we present ReVision, a retinal foundation model that enables efficient deployment across diverse downstream applications.

Our systematic evaluation across 27 benchmarks establishes ReVision's deployment efficiency. First, ReVision demonstrates exceptional zero-shot diagnostic capability, achieving an average AUROC of 0.946 across 12 public benchmarks and 0.952 on three real-world multi-disease cohorts without any task-specific training. This performance matches or exceeds fully supervised models, effectively eliminating the need for extensive local data annotation. Zero-shot pathology localization further provides interpretable visual evidence, addressing a key barrier to clinical trust. When down-stream adaptation is feasible, ReVision exhibits exceptional efficiency in both computational demands and data requirements. Linear probing on frozen embeddings outperforms full fine-tuning of competing models while requiring 10,000-fold fewer trainable parameters. Training with merely 10\% of available data surpasses the performance of RETFound fully fine-tuned on complete datasets, demonstrating remarkable sample efficiency. Furthermore, ReVision's learned representations transfer effectively across clinical sites, imaging devices, patient populations, imaging modalities, and novel oculomics tasks. Together, these properties address practical barriers that have hindered FM deployment in resource-constrained settings.

The deployment efficiency of ReVision goes beyond traditional diagnostic metrics, offering meaningful assistance in real-world clinical practice. Our prospective reader study involving 33 ophthalmologists demonstrates that ReVision's zero-shot assistance improves diagnostic accuracy by 14.8\% across all experience levels, with the model functioning as a collaborative tool that augments rather than replaces clinical judgment. Through diagnostic confidence analysis, we demonstrate that ReVision enhances clinicians' confidence in routine cases and triggers alert signals in challenging or divergent scenarios. Additionally, Likert scale ratings of AI-assisted information highlight clinicians' strong focus on the interpretability of the results, as opposed to simply the outcome itself. Detailed behavioral analysis further reveals complex interaction patterns, showing how clinicians appropriately resist incorrect AI suggestions in critical cases while leveraging AI guidance to correct potential diagnostic errors, reflecting ideal human-AI collaboration. However, in some cases, the potential misguidance and risks introduced by AI, along with clinicians' susceptibility to blind reliance, have also been observed, raising warnings for AI-assisted workflows. Rather than merely presenting the improvements in diagnostic accuracy, the objective analysis of behavioral patterns, along with the presentation of clinicians' subjective feedback, offers a more comprehensive perspective on the human-AI collaboration workflow, highlighting real-world challenges and providing valuable insights for future deployment.

Despite these promising results, several limitations warrant careful consideration. First, while ReVision's text-prompted zero-shot classification demonstrates strong performance on threshold-free metrics (AUROC, AUPR), clinical deployment would benefit from threshold calibration using local validation sets, as different clinical scenarios demand different operating points. Second, the current ReVision only supports CFPs. Although the learned representations naturally extend to other \textit{en face} imaging modalities, as demonstrated in our cross-modality experiments, adaptation to volumetric imaging modalities like OCT would require architectural modifications and additional training data. Third, our training data, while extensive, predominantly reflects patterns observed in East Asian populations. Although evaluation on globally diverse benchmarks demonstrates encouraging generalization, population-specific validation remains essential before widespread international deployment\cite{tham2025building}. Finally, privacy considerations merit careful attention, as vision-language models trained on large patient populations may potentially encode sensitive information, particularly relevant for generative applications\cite{chen2024generative}.

Future research directions offer exciting opportunities to enhance clinical impact and expand capabilities. Most critically, prospective randomized controlled trials (RCT) in real-world clinical settings are essential to rigorously validate ReVision's effectiveness in improving patient outcomes, cost-effectiveness, and seamless integration into clinical workflows\cite{eyefm}. Beyond clinical validation, several technical extensions could substantially expand ReVision's utility. Incorporating longitudinal patient data could enable prediction of disease progression trajectories and treatment responses, transforming static diagnosis into dynamic prognosis and personalized medicine\cite{dai2024deep}. The unified vision-language embedding space provides a natural foundation for multi-modality integration, potentially aligning OCT, fluorescein angiography, ultra-widefield imaging, and other modalities through shared clinical descriptions\cite{shi2024eyeclip}. Integration with large language models could enable comprehensive diagnostic report generation and interactive clinical decision support systems\cite{singhal2025toward, eyefm, li2023llava}, facilitating more natural and intuitive human-AI collaboration in clinical practice.

In conclusion, ReVision demonstrates that large-scale telemedicine archives provide an effective source for building deployable clinical FM.
The native intelligence embedded in real-world clinical documentation provides more valuable insights than any carefully curated research dataset. The telemedicine archives that powered ReVision represent just one example. Such archives exist across healthcare systems worldwide, representing an underutilized resource for developing FMs that capture diagnostic reasoning as it occurs in clinical practice.

%% file: method.tex
\section*{Methods}

\subsection*{Data curation for ReVision pretraining}

The scale and quality of data are vital for self-supervised learning. 
We curated TM500k from a telemedicine program coordinated by Beijing Tongren Hospital, a national tertiary ophthalmology center. Through an online consultation platform, 172 primary hospitals across China uploaded queries containing demographic data, visual acuity, chief complaints, medical histories, and ophthalmic images. Senior ophthalmologists in the reader center provided diagnostic reports including findings, impressions, and clinical advice.

From an initial collection of 918,255 images across 419,412 patient visits, we applied a multi-stage curation pipeline. We trained an ensemble of classification models to exclude non-CFP images (n=275,826), removed images with suboptimal quality identified through textual cues in reports (n=36,248), excluded images lacking corresponding findings and impressions (n=40,662), and retained only one representative CFP per eye per patient to reduce redundancy (n=79,539). Data from 162 hospitals were used for pretraining, with 10 hospitals reserved for downstream evaluation (TMMC). This process yielded 485,980 CFP images paired with diagnostic reports (Supplementary Fig. 1).

Each report comprises three components: medical history (prior illnesses, systemic diseases from patient self-report), findings (objective observations of fundus structures and abnormalities), and impression (diagnostic summary). For bilateral images, we segmented reports by eye-specific keywords; when absent, findings were assigned to both eyes. Original Chinese reports were translated to English using GPT-4o to enable broader applicability. Detailed data curation pipeline is described in Supplementary Materials.

\subsection*{Data for downstream validation}

We established 27 benchmarks spanning seven countries to evaluate ReVision across diverse clinical tasks. 

For ocular disease detection and classification tasks, we employed 12 benchmarks. Diabetic retinopathy (DR) was evaluated on IDRID\cite{IDRID} (India), MESSIDOR-2\cite{MESSIDOR2} (France), and APTOS-2019\cite{APTOS} (India). Glaucoma detection was evaluated on REFUGE\cite{REFUGE} (China) and AIROGS\cite{AIROGS} (USA). Myopia pathology detection was evaluated on MMAC\cite{MMAC} (China). Age-related macular degeneration (AMD) detection was evaluated on ADAM\cite{ADAM} (China) and MMC-AMD\cite{MMCAMD} (China). Multi-disease single-label classification was evaluated on JSIEC\cite{JSIEC} (China) and Retina\cite{Retina} (source not recorded). Multi-disease multi-label classification (one image can belong to multiple categories) was evaluated on ODIR\cite{ODIR} (China) and RFMiD\cite{RFMiD} (India).

For dense prediction tasks, focusing on the localization of various lesions and pathological changes, we employed 3 benchmarks.
For diabetic retinopathy lesions (hard exudates, soft exudates, hemorrhages), we integrated IDRID\cite{IDRID} (India), FGADR\cite{FGADR} (UAE), MAPLES-DR\cite{MAPLES-DR} (France), and E-ophtha\cite{eophtha} (France). For pathological myopia lesions (CNV, fuchs spot, lacquer cracks, atrophy, detachment), we integrated MMAC\cite{MMAC} (China) and PALM\cite{PALM} (China). For AMD lesions (drusen, exudates, hemorrhages, scars), we used ADAM\cite{ADAM} (China).

For cross-domain and cross-modality tasks, we employed 4 benchmarks. We conducted evaluations across datasets that differ from the pretraining data in age distribution, imaging devices, and modalities, including SZEH\cite{ROP} (China; retinopathy of prematurity grading), mBRSET\cite{mBRSET} (Brazil, portable handheld cameras), MPOS\cite{MPOS} (China; fundus fluorescein angiography), and OUWF\cite{OUWF} (China, ultra-widefield imaging). 

For oculomics tasks, we employed 5 benchmarks. We used UK Biobank\cite{ukbb} retinal images from 51,542 participants after quality filtering, identifying incident ischemic stroke, myocardial infarction, and heart failure within 5 years of baseline imaging. For carotid intima-media thickness prediction, we used CF-CIMT (China). For systemic biomarker estimation, we analyzed 17 biomarkers from 37,475 UK Biobank participants. The Birmingham assessment center served as the external test set.

In addition to public benchmarks, we employed 3 real-world clinical cohorts for zero-shot evaluation. The Telemedicine Multi-disease Multi-label Cohort (TMMC; n=35,940 images, 30 disease categories) was held out from ReVision's pretraining data, with labels extracted from diagnostic reports. The Shanghai Ruijin Multi-class Cohort (SRMC; n=1,461 images, 13 conditions) comprises CFPs with disease annotations confirmed by corresponding OCT examinations. The Handan Multi-view Cohort (HMVC; n=5,362 images, 2,687 eyes, 5 conditions) provides paired macula-centered and optic disc-centered views from an opportunistic screening program. The predictions of both views per eye is averaged to generate eye-level prediction.

Detailed dataset specifications, imaging devices, and data splits are provided in Supplementary Material and Supplementary Tables S1–S4.

\subsection*{Image preprocessing}
For CFPs used in both pre-training and downstream transfer tasks, we applied preprocessing using AutoMorph\cite{automorph} to extract the retinal region while removing irrelevant background artifacts, followed by resizing to a square shape. For FFA images, we removed background areas based on pixel intensity thresholds, preserving the retinal region and padding the images to a square shape while maintaining their original aspect ratios. For UWF fundus images, preprocessing primarily involved border padding to convert the original images into standardized square formats.

\subsection*{Visual-language pretraining}

ReVision employs a vision-language contrastive learning framework consisting of an image encoder and a text encoder. The image encoder utilizes a Vision Transformer (ViT-Large/14) architecture\cite{vit2021}, processing 336×336 pixel images divided into 14×14 pixel patches. The ViT backbone comprises 24 transformer layers with 16 attention heads and an embedding dimension of 1024, with a [CLS] token capturing global image representations. The text encoder adopts a 12-layer Transformer architecture to process diagnostic reports (maximum 128 tokens), with an [EOT] token representing global textual information. Both encoders employ linear projectors to map their outputs into a shared 768-dimensional embedding space. Model parameters are initialized from OpenAI-CLIP ViT-Large/14@336px\cite{radford2021clip}, pretrained on 400 million internet-sourced image-text pairs.

\textbf{Contrastive learning objective.} For a batch of $N$ image-text pairs, we denote image embeddings as $\{\mathbf{u}_i\}_{i=1}^N$ and text embeddings as $\{\mathbf{v}_i\}_{i=1}^N$. The normalized cosine similarity between the $i$-th image and $j$-th text is:
\begin{equation}
s_{ij} = \frac{\mathbf{u}_i^\top \mathbf{v}_j}{\|\mathbf{u}_i\| \cdot \|\mathbf{v}_j\|}.
\end{equation}
Following CLIP\cite{radford2021clip}, we optimize bidirectional contrastive objectives that maximize similarity between matched pairs while minimizing similarity between unmatched pairs:
\begin{equation}
\mathcal{L}_{\text{i2t}} = -\frac{1}{N}\sum_{i=1}^N \log \frac{\exp(s_{ii}/\tau)}{\sum_{j=1}^N \exp(s_{ij}/\tau)},
\end{equation}
\begin{equation}
\mathcal{L}_{\text{t2i}} = -\frac{1}{N}\sum_{j=1}^N \log \frac{\exp(s_{jj}/\tau)}{\sum_{i=1}^N \exp(s_{ij}/\tau)},
\end{equation}
\begin{equation}
\mathcal{L}_{\text{CLIP}} = \frac{1}{2}(\mathcal{L}_{\text{i2t}} + \mathcal{L}_{\text{t2i}}),
\end{equation}
where $\tau$ is a learnable temperature parameter controlling the concentration of the distribution. 

To enhance semantic consistency between modalities, we introduce an alignment loss that encourages structural similarity between image and text embedding spaces:
\begin{equation}
\mathcal{L}_{\text{align}} = \frac{1}{N^2}\sum_{i=1}^N\sum_{j=1}^N (s_{ij}^{\text{img}} - s_{ij}^{\text{txt}})^2,
\end{equation}
where $s_{ij}^{\text{img}}$ and $s_{ij}^{\text{txt}}$ denote the cosine similarities between the $i$-th and $j$-th samples in image and text embedding spaces respectively, computed analogously to Equation~(1). The final training objective combines these losses:
\begin{equation}
\mathcal{L}_{\text{base}} = \mathcal{L}_{\text{CLIP}} + \lambda_{\text{align}}\mathcal{L}_{\text{align}},
\end{equation}
where $\lambda_{\text{align}}=1$ in our implementation.

\textbf{Demographic-augmented training for oculomics.} For oculomics tasks predicting systemic conditions from retinal images, we train an enhanced variant incorporating auxiliary demographic prediction objectives. These tasks encourage the model to capture subtle indicators of overall health status beyond explicit retinal pathologies. Using the image encoder's penultimate features $\mathbf{f}_i$ (before the projection layer), we add:
\begin{equation}
\mathcal{L}_{\text{sex}} = -\frac{1}{N}\sum_{i=1}^N \left[y_i\log\sigma(\mathbf{w}_s^\top\mathbf{f}_i) + (1-y_i)\log(1-\sigma(\mathbf{w}_s^\top\mathbf{f}_i))\right],
\end{equation}
\begin{equation}
\mathcal{L}_{\text{age}} = \frac{1}{N}\sum_{i=1}^N (\mathbf{w}_a^\top\mathbf{f}_i - a_i)^2,
\end{equation}
where $\mathbf{w}_s$ and $\mathbf{w}_a$ are learnable weight vectors, $\sigma(\cdot)$ denotes the sigmoid function, and $y_i \in \{0,1\}$ and $a_i$ represent ground-truth sex and age values respectively. The demographic-augmented training objective becomes:
\begin{equation}
\mathcal{L}_{\text{demo}} = \mathcal{L}_{\text{CLIP}} + \lambda_{\text{align}}\mathcal{L}_{\text{align}} + \lambda_{\text{age}}\mathcal{L}_{\text{age}} + \lambda_{\text{sex}}\mathcal{L}_{\text{sex}},
\end{equation}
where $\lambda_{\text{age}}=1$ and $\lambda_{\text{sex}}=0.1$ to balance the contribution of auxiliary tasks.

\textbf{Training configuration.} Both model variants were optimized using AdamW\cite{loshchilov2019adamw} with learning rate $3\times10^{-5}$, weight decay $1\times10^{-3}$, $\beta=(0.9, 0.98)$, and $\epsilon=1\times10^{-6}$. Training proceeded for 3 epochs (486,400 steps) on TM500k with batch size 512. We employed cosine learning rate decay with 200-step linear warmup and exponential moving average (EMA, decay=0.995) for parameter stabilization. Standard augmentations included random cropping, color jittering, horizontal flipping, and Cutout. Training was distributed across 2 NVIDIA RTX A6000 GPUs using mixed precision (FP16) and gradient checkpointing for memory efficiency.

\subsection*{Implementation of zero-shot prediction}
To evaluate ReVision's zero-shot capabilities, we leveraged the model's inherent vision-language alignment achieved during pretraining. Each class was associated with several simple text prompts consisting of the class name. For example, we used "normal fundus" and "no abnormalities" for the normal class; we used "glaucoma" and "suspected glaucoma" for the glaucoma class. The averaged text embeddings of the prompts are used as the prompt embedding of the class. The output classification probabilities were computed by cosine similarity between image embedding and prompt embeddings. To build unambiguous zero-shot benchmarks, we trimmed the original datasets. We transformed DR grading datasets, \textit{i.e.}, APTOS-2019, MESSIDOR-2, IDRID, and mBRSET, to binary DR detection benchmarks. For MMC-AMD, we merged the "PCV" class (sub-type of wet-AMD) into wet-AMD to avoid ambiguity. For Retina, RFMID, and ODIR, we ignored the "other diseases" class. For JSIEC with 39 classes, we merge sub-classes "DR1", "DR2", and "DR3", into "DR"; we merged "rhegmatogenous RD" and "dragged disc" into "retinal detachment"; and we ignored "blur fundus with suspected PDR" and "blur fundus without PDR" as they are the sub-classes of DR and Normal.  

To further explore the interpretability of ReVision’s diagnostic predictions and its capability for zero-shot lesion localization, we adopted the ClearCLIP\cite{lan2024clearclip} technique to generate spatially resolved attention maps guided by text prompts. For a given input image, patch-level features were extracted from the frozen visual encoder and normalized. Each clinical concept, either a disease category (\textit{e.g.}, "diabetic retinopathy") or a lesion type (\textit{e.g.}, "hard exudates"), was encoded into a text embedding via the text encoder. Cosine similarities between each image patch and the text embedding were then computed via element-wise multiplication followed by summation across the feature dimension. The resulting similarity scores were reshaped into a 2D spatial grid and upsampled to match the original image resolution, yielding a high-resolution activation map that highlights regions most aligned with the semantic meaning of the prompt. For comparative evaluation against baseline zero-shot methods, we employed the image size for pre-training of each model, \textit{i.e.}, 224$\times$224 for BiomedCLIP and RetiZero, 336$\times$336 for OpenAI-CLIP(Large@336px), and 512$\times$512 for FLIAR. It is worth noting that in zero-shot inference without fine-tuning, uniformly standardizing the image resolution across different models is not fair, as this would lead to inconsistency between zero-shot inference and pre-training, thereby resulting in degraded performance.

\subsection*{Implementation of supervised downstream Learning}
To enable downstream adaptation, we attached task-specific prediction heads to the pre-trained image encoder of ReVision. For classification tasks, the image encoder processed input images at a resolution of 336 × 336 to extract feature embeddings, which were subsequently passed through a linear classification head to generate probability outputs. For ocular diseases, we utilized the feature embedding after the linear projector. For oculomics, we utilized the pooled feature before the projector. We explored two adaptation strategies: linear probing (LP) and full fine-tuning (FT). In LP, the encoder was frozen and only the classification head was trained. The model was optimized with the batch size of 16 for 20 epochs using AdamW optimizer with learning rate = $5\times10^{-4}$, weight decay = $1\times10^{-2}$, and $\beta$ = (0.9, 0.999). In FT, both the encoder and the classification head were updated jointly to achieve optimal task-specific performance. The training setup is the same, except that the encoder parameters were updated with a lower learning rate ($5\times10^{-7}$ for ocular disease, and $5\times10^{-6}$ for oculomics) and $\beta$ = (0.9, 0.98). The checkpoint with the best performance (AUROC + 0.5×AUPR) on the validation set was saved for testing. Standard image augmentation techniques were applied to avoid over-fitting, including random cropping, color jittering, horizontal flipping, Cutout, and Mixup.

For supervised lesion segmentation tasks, input images were resized to 448$\times$448 and encoded into multi-scale feature representations by the image encoder. Features from layers 6, 12, 18, and 24 were fed into a UNETR\cite{unetr} segmentation head, enabling end-to-end prediction of dense lesion masks. During adaptation, the segmentation head was trained while the encoder remained frozen. We used a batch size of 32 and optimized with AdamW (learning rate = $1\times10^{-3}$) for 100 epochs using a combination of Dice loss and focal loss.

For comparative evaluation against baseline methods, we employed the same image size (336$\times$336) by default for all models for fair comparison. For RETFound and VisionFM, we utilized their official fine-tuning code implementations. Downstream training was conducted on NVIDIA GeForce RTX 3090/4090 GPUs. 

\subsection*{Reader study}

The reader study employed a structured two-phase diagnostic protocol designed to evaluate both baseline clinical performance and the incremental value of AI assistance. The image data was randomly selected from outpatient encounters at Beijing Tongren Hospital between January and April 2025 to encompass 21 distinct ocular conditions, providing a comprehensive representation of common and rare retinal pathologies encountered in clinical practice. Ground truth diagnoses were established by chief ophthalmologists through comprehensive clinical evaluation incorporating patient history, chief complaints, fundoscopy, OCT, FFA, etc. We excluded images with insufficient quality for reliable grading and cases presenting multiple concurrent pathologies to ensure clarity during the reader study. The final dataset comprised 154 color fundus photographs. All participants evaluated the complete image set to ensure robust statistical comparability while maintaining a manageable workload.

We recruited 33 ophthalmologists representing the full spectrum of clinical expertise according to the Chinese medical hierarchy system. The cohort comprised trainees (n=7, graduate students with less than 3 years of clinical experience), juniors (n=10, attending physicians, 3-10 years of experience), seniors (n=7, associate chief physicians, 10-15 years of experience), and experts (n=9,  chief physicians, 11-25 years of experience).
Participants were recruited from 12 institutions spanning diverse geographical regions in China. This institutional diversity ensured representation of tertiary centers.

The reader study was conducted through a secure web-based platform specifically developed for this study, enabling standardized data collection across geographically distributed participants (Supplementary Figure 6). Each participant completed the evaluation through the following standardized workflow. There are two consecutive stages for each image.

\textit{Stage 1: Independent clinical assessment.} Participants initially performed an unassisted diagnostic evaluation. During this stage, clinicians examined the image without any AI support and recorded their primary diagnosis from the predefined list of 21 ocular conditions. Participants also documented their diagnostic confidence using a five-point Likert scale (1 = very uncertain, 2 = uncertain, 3 = neutral, 4 = confident, 5 = very confident).

\textit{Stage 2: AI-assisted diagnosis.}
Following the initial assessment, the platform presented ReVision's diagnostic assistance comprising three complementary components: (1) the top-5 most probable diseases ranked by the model's zero-shot prediction confidence, (2) the top-5 most likely retinal lesions or abnormalities identified, and (3) class activation mapping (CAM) heatmaps highlighting regions of pathological significance within the fundus image. After reviewing this AI-generated information, participants were given the opportunity to revise their initial diagnosis if deemed appropriate. They then recorded their final diagnostic decision and reassessed their confidence level using the same five-point scale.
For each case, participants evaluated the clinical utility of each AI assistance component individually. They rated the practical value of the top-5 disease predictions, top-5 lesion identifications, and CAM visualization in supporting their diagnostic decision-making process.

To ensure data integrity, the platform incorporated several quality control mechanisms. These included mandatory completion of all assessment fields before case progression and prevention of retroactive modifications to previous cases. Upon completing all 154 cases, participants completed a post-study questionnaire assessing their overall experience with the AI-assisted workflow. This questionnaire captured subjective perceptions regarding the system's clinical value.

\subsection*{Evaluation and statistical analysis}
Image-level classification performance was evaluated by the area under the receiver operating characteristic curve (AUROC) and precision-recall curve (AUPR). These general metrics do not require selecting a threshold for the model’s probability outputs, which may need tuning for different data and models. For binary classification, AUROC and AUPR were calculated in a binary setting. For multi-class classification, we computed these metrics for each class and averaged them (`macro' setting). In addition, we employed accuracy (ACC), sensitivity, specificity, precision, and F1-score. To convert the probability values output by the model into categories, we optimized decision thresholds on the validation set of each public dataset and applied them to the test set. Segmentation performance was evaluated by Dice Similarity Coefficient (DSC) and Intersection over Union (IoU) scores, which quantifies the overlap between predicted segmentation masks and ground truth annotations. For zero-shot localization, we searched for the best threshold on the predicted heatmaps to obtain the theoretical optimal segmentation masks and reported the corresponding DSC and IoU scores \cite{bercea2025evaluating, defard2021padim}. In addition, we adopted the per-region overlap area under the curve (PRO \cite{bergmann2020uninformed}) for zero-shot localization, which evaluates localization performance by integrating the curve of per-region overlap versus false positive rate (FPR) within the low-FPR regime ($\mathrm{FPR} \in [0, 0.3]$). Regression performance (biomarker prediction) was evaluated by Pearson's correlation and coefficient of determination ($R^2$). For supervised classification and segmentation tasks that require downstream training, we trained the model five times with different randomizations. We calculated the mean and standard deviation of performance across the five runs, and the 95\% CI deviation is 1.96 $\times$ standard deviation /$\sqrt{5}$. P-value was calculated using two-sided student t-tests to check for statistical significance. For zero-shot tasks that directly employ the pre-trained model and regression tasks, we employed non-parametric bootstrapping with 2,000 resamples to derive 95\% CIs. P-values were calculated as twice the proportion of bootstrap samples where the better model performed worse than the other. Performance of all downstream datasets is presented in Supplementary Table S5-S12.

\subsection*{Ethics Statement}
The study was conducted according to the tenets of the Declaration of Helsinki. This study and the telemedicine data were approved by the institutional review board of Beijing Tongren Hospital of Capital Medical University (TREC2024-KYS428). Prior to enrollment in the telemedicine program, all participating hospitals were notified that the uploaded images and the diagnostic reports generated by Tongren Hospital could be used anonymously for educational and research purposes. Because the study was a retrospective review and analysis of fully anonymized images, the medical ethics committee declared it exempt from informed consent.  SRMC was approved by the institutional review board of Ruijin Hospital, Shanghai Jiao Tong University School of Medicine (Ethics 2020 No.183). HMVC was approved by the institutional review board of  Handan City Eye Hospital (2022003). UK Biobank is used under Application Number 106544.

\section*{Data Availability}
The data used for pre-training is not directly public available as it contains sensitive information. The deidentified data can be shared only for academic purposes and will require a formal material transfer agreement. Requests can be submitted by emailing N. Wang and H. Li. SRMC can be requested by emailing F. Chen and H. Yao. HMVC can be requested by emailing N. Wang. Other downstream datasets are directly available or available upon simple request. The links of the resources are presented as followed: IDRID (\url{https://ieee-dataport.org/open-access/indian-diabetic-retinopathy-image-dataset-idrid}), MESSIDOR-2 (\url{https://www.adcis.net/en/third-party/messidor2/}), APTOS-2019 (\url{https://www.kaggle.com/competitions/aptos2019-blindness-detection/data}), JSIEC (\url{https://zenodo.org/record/3477553}), Retina (\url{https://www.kaggle.com/datasets/jr2ngb/cataractdataset}), (above datasets are also available from RETFound), REFUGE (\url{http://hdmilab.cn/ichallenge}, \url{https://refuge.grand-challenge.org/}), AIROGS-v2 (\url{https://www.kaggle.com/datasets/deathtrooper/glaucoma-dataset-eyepacs-airogs-light-v2}), MMAC (\url{https://zenodo.org/records/11025749}), MMC-AMD (\url{https://github.com/li-xirong/mmc-amd}), ADAM (\url{http://hdmilab.cn/ichallenge}, \url{https://amd.grand-challenge.org/}), RFMID (\url{https://ieee-dataport.org/open-access/retinal-fundus-multi-disease-image-dataset-rfmid}), ODIR (\url{https://github.com/nkicsl/OIA-ODIR}), FIVES (\url{https://doi.org/10.6084/m9.figshare.19688169.v1}), SZEH (\url{https://doi.org/10.6084/m9.figshare.25514449}), mBRSET (\url{https://physionet.org/content/mbrset/1.0/}), MOPS (\url{https://github.com/whq-xxh/FFA-Synthesis}), OUWF (\url{https://doi.org/10.6084/m9.figshare.26936446}), CF-CIMT (\url{https://doi.org/10.6084/m9.figshare.27907056}).   UK Biobank, a publicly accessible database, is available to researchers through an application via \url{https://www.ukbiobank.ac.uk/register-apply/}.

\section*{Code and Model Availability}
Code and ReVision model weights of different model sizes will be public available upon publication.

\section*{Acknowledgements}
This study was supported by the National Natural Science Foundation of China (NSFC) (Grant No. 82572313, Grant No. 62506036, and Grand No. U22A2051), Hebei Natural Science Foundation (Grant No. F2025105040), the Excellent Young Scientists Fund of the National Natural Science Foundation of China (Grant No. 82422018), Beijing Municipal Public Welfare Development and Reform Pilot Project for Medical Research Institutes (PWD\&RPP-MRI, JYY2023-6), Sanming Project of Medicine in Shenzhen (Grant No. SZSM202411009), Beijing Natural Science Foundation (Grant No. F251001), and National key Research and Development Program of China (Grant No. 2022YFC2405200). Part of this research has been conducted using the UK Biobank Resource under application number 106544.

\section*{Competing interests}
The authors declare no competing interests.